\documentclass{statsoc}

\usepackage[a4paper]{geometry}
\usepackage{graphicx}
\usepackage[textwidth=8em,textsize=small]{todonotes}
\usepackage{amsmath,amssymb}
\usepackage{natbib}

\usepackage[greek,UKenglish]{babel}

\usepackage[colorlinks=true, allcolors=blue]{hyperref}
\usepackage{tcolorbox}

\usepackage{ragged2e} 

\usepackage[autostyle=false, style=english]{csquotes}
\MakeOuterQuote{"}

\allowdisplaybreaks 

\usepackage{fancyvrb} 

\usepackage[title]{appendix}

\usepackage{algpseudocode}
\usepackage{algorithm}
\usepackage{booktabs}
\usepackage{threeparttable}
\usepackage{multirow}

\newcommand{\argmax}{\arg\,\max} 
\DeclareMathOperator{\diag}{diag}

\DeclareMathOperator{\Bin}{Bin}
\DeclareMathOperator{\Mult}{Mult}
\DeclareMathOperator{\InvGamma}{Inv \, Gamma}

\DeclareMathOperator{\Var}{Var}

\title[Measuring diachronic sense change]{Measuring diachronic sense change: new models and Monte Carlo methods for Bayesian inference}
\author{Schyan Zafar}
\address{Department of Statistics, University of Oxford}
\email{schyan.zafar@jesus.ox.ac.uk}
\author[Zafar and Nicholls]{Geoff Nicholls}
\address{Department of Statistics, University of Oxford}
\email{nicholls@stats.ox.ac.uk}

\begin{document}

\begin{abstract}
In a bag-of-words model, the \textit{senses} of a word with multiple meanings, e.g. "bank" (used either in a river-bank or an institution sense), are represented as probability distributions over context words, and sense prevalence is represented as a probability distribution over senses. Both of these may change with time. Modelling and measuring this kind of sense change is challenging due to the typically high-dimensional parameter space and sparse datasets. A recently published corpus of ancient Greek texts contains expert-annotated sense labels for selected target words. Automatic sense-annotation for the word "kosmos" (meaning decoration, order or world) has been used as a test case in recent work with related generative models and Monte Carlo methods. We adapt an existing generative sense change model to develop a simpler model for the main effects of sense and time, and give MCMC methods for Bayesian inference on all these models that are more efficient than existing methods. We carry out automatic sense-annotation of snippets containing "kosmos" using our model, and measure the time-evolution of its three senses and their prevalence. As far as we are aware, ours is the first analysis of this data, within the class of generative models we consider, that quantifies uncertainty and returns credible sets for evolving sense prevalence in good agreement with those given by expert annotation.\\

\textit{Keywords:} Bayesian inference; diachronic lexical semantics; Markov Chain Monte Carlo methods; natural language processing (NLP); sense change model

\end{abstract}

\section{Introduction}
\label{sec:introduction}

As a natural language evolves, the meanings of words within the language change. The field of diachronic lexical semantics is concerned with how word meanings change over time. Words with multiple meanings or \textit{senses}, and their time-evolution, are of considerable interest within the field. Examples of such words include "mouse" (meaning a rodent or a computer pointing device) and "bank" (meaning a river-bank or a financial institution). Statistical models of diachronic sense change are useful for lexicographic and linguistic research as well as for downstream applications in many natural language processing (NLP) tasks such as information retrieval.

For a word with multiple senses, the intended sense is usually apparent from the context. For example, the different senses of "bank" are obvious in the text snippets "deposited £500 in his bank account" and "plants growing on the bank of the Indus river". We expect certain context words to be used more often than others depending on the intended sense of the target word. In the "bank" example for instance, context words such as "money" or "account" are more likely when "bank" is used in the financial institution sense, whereas context words such as "river" or "stream" are more likely in the river-bank sense. We may also expect changes in the relative frequency of context words over time. E.g. for the financial institution sense of "bank", the context word "specie" (coin) is more likely to be used up to the early 20\textsuperscript{th} century whereas the context word "card" is more likely to be used later on. The prevalence of a sense itself may change over time, e.g. the pointing device sense of "mouse" increased in prevalence over the later half of the 20\textsuperscript{th} century. 
We are interested in a model that captures all of these features.

In this paper we analyse (as a test case) the sense change for the ancient Greek word "kosmos", meaning decoration, order or world, and quantify the uncertainty in these sense change estimates. Obtaining usefully narrow credible intervals with good coverage is no easy task and, viewed from a statistical perspective, this is our main contribution to the field. We achieve this by careful statistical modelling of the data. We develop a new model of Diachronic Sense Change (DiSC) by adapting the Sense ChANge (SCAN) and Genre-Aware Semantic Change (GASC) models of \citet{frermann2016bayesian} and \citet{GASC_2019arXiv190305587P} respectively. Under this modelling framework, target word senses are represented as probability distributions over context words, sense prevalence is represented as a probability distribution over senses, and both sense and sense prevalence have temporal dependence. Our DiSC model has significantly fewer parameters than SCAN/GASC. However, there is no evidence for any loss of goodness-of-fit, and our model is significantly easier to analyse. Both aspects are important when the word of interest occurs infrequently in a very large fixed corpus of surrounding text as is the case for "kosmos" in the ancient Greek data. We found that we could only give a reliable and well-calibrated fit for SCAN/GASC when the data were strongly informative of the parameters. We could not find any variant of MCMC that could reliably fit SCAN/GASC to the "kosmos" data even for point estimation of parameters, let alone uncertainty quantification. On the other hand, we were able to fit DiSC to the data, quantify uncertainty, and get a well-calibrated match to the expert sense-annotation. Moreover, in experiments on synthetic data generated according to the SCAN model itself, DiSC scores at least as well as SCAN on sense-labelling, as measured by Brier scores. We attribute this to a favourable bias-variance tradeoff.

These models are related to topic models, though there is no direct correspondence. Variational methods are widely used to fit topic models and can be efficient for inferring posterior means. However, variational methods are less reliable for quantifying uncertainty, and in particular tend to underestimate variance. 
Markov Chain Monte Carlo (MCMC) methods are particularly challenging for the models under discussion. However, where MCMC methods can be shown to converge, they at least give asymptotically exact posterior summaries. We present a relatively efficient method for fitting these models in such cases, where we marginalise the posterior by summing over the discrete sense labels, and use gradient-based MCMC schemes to target the remaining continuous distributions. We use the occurrences of "bank" in an English text corpus, as a simple illustrative example where all analyses are possible, to compare our sampler against existing MCMC samplers for these models. We compare the models' predictive performance on held-out sense labels for "bank" and for synthetic datasets. We then analyse "kosmos" using our new model and MCMC sampler.

The rest of this paper is structured as follows. In Section~\ref{sec:related_work} we review the existing approaches for modelling diachronic semantic change and parameter inference within the NLP literature. In Section~\ref{sec:data} we describe the "bank" and "kosmos" datasets, respectively from the English and ancient Greek corpora. In Section~\ref{sec:the_model} we present our new DiSC model and highlight the differences with the existing SCAN and GASC models. In Section~\ref{sec:inference} we describe the existing MCMC samplers targeting the posterior, and present our MCMC scheme. Section~\ref{sec:experiment_samplers} compares the performance of the MCMC samplers on the SCAN model for "bank". Sections~\ref{sec:experiment_bank}--\ref{sec:experiment_kosmos} look at the application of the models to the English, synthetic and ancient Greek datasets. Section~\ref{sec:conclusion} concludes with a discussion of limitations and possible future research. Some further technical details and results are given in the appendices, and R scripts for data extraction and model-fitting are uploaded to GitHub.

\section{Related work}
\label{sec:related_work}

The problem of modelling diachronic semantic change has been approached in several different ways within the field of NLP, and detailed overviews of the literature are given by \citet{2018arXiv181106278T} and \citet{2018arXiv180109872T}. Broadly, the approaches can be categorised into three groups: topic-based models, graph-based models and embedding models.

The topic model first introduced by \citet{blei2003latent} is a generative model called Latent Dirichlet Allocation (LDA). Under LDA, a document of given length is generated by sampling a topic, and then a word given the topic, at each position in the document. LDA is a simple bag-of-words model that captures some basic ideas of meaning via the word-topic associations. The model uses Dirichlet priors for probability distributions over topics and words, and variational inference is commonly used for parameter estimation. \citet{griffiths2004finding} give a collapsed Gibbs sampler targeting the marginal posterior integrated over the continuous parameters, which can be used for asymptotically exact inference.

LDA was extended by \citet{Blei:2006:DTM:1143844.1143859} to give a dynamic topic model which additionally captures the time-evolution of topics. The dynamic topic model uses logistic normal priors since these are straightforward to model as a time series, as opposed to the parameters of the Dirichlet priors under LDA. Variational inference is used for the parameters, but alternative sampling methods have been proposed in the literature. These include the blocked Gibbs sampler based on auxiliary uniform variables given by \citet{mimno2008gibbs} and a strategy based on auxiliary Polya-Gamma \citep{PolyaGamma_2012arXiv1205.0310P} variables given by \citet{chen2013scalable}.

The dynamic topic model was adapted by \citet{frermann2016bayesian} to explicitly capture the meanings or senses of a given target word (as opposed to topics in documents) and their time-evolution in the SCAN model. The main distinction between the two is that a topic model has an independent topic underlying each context word, whereas in SCAN all context words for a single usage of the target word share the same sense. The logistic normal priors, defined separately in each time period, are connected to their temporal neighbours via an intrinsic Gaussian Markov Random Field \citep{rue2005gaussian}, which enables modelling the change in adjacent parameters without requiring a global mean. GASC, an extension to SCAN given by \citet{GASC_2019arXiv190305587P}, additionally allows the prevalence of each sense to vary according to the genre of the text in which the target word is used. Both authors use the Gibbs sampler of \citet{mimno2008gibbs} for inferring the model parameters.

A distinct graph-based approach to this problem is given by \citet{2014arXiv1405.4392M, mitra2015automatic} who use a semantic network model in which words are represented as nodes, and edges between nodes denote word co-occurrence in a sentence. The senses of a word are clustered separately for two different time periods, and then compared across the time periods to identify sense changes as well as sense births, deaths, mergers and splits. A similar approach is used by \citet{tahmasebi2017finding} who cluster senses separately for each time period and track the clusters through time. 

Word embeddings are techniques for mapping words onto low-dimensional real vector spaces. In recent years, the neural-network-based Word2vec models developed at Google by \citet{2013arXiv1301.3781M} have emerged as the most popular word embedding models, although Stanford's GloVe model developed by \citet{pennington2014glove} is a popular alternative based on a factorisation of the global co-occurrence matrix. Skip-gram is the more popular of the two Word2vec models (the other being CBOW) and has been used to capture many semantic word relationships \citep{NIPS2013_5021} and linguistic regularities \citep{mikolov-etal-2013-linguistic}, but the original model only uses one vector representation for each word and hence does not allow for multiple senses. The original Skip-gram has been extended in several ways to capture multiple senses per word, e.g. the Adaptive Skip-gram model given by \citet{bartunov2016breaking} and the loss driven multi-sense identification (LDMI) model given by \citet{2019arXiv190406725M}. A comprehensive review of embedding techniques used for word sense representation is given by \citet{CamachoCollados2018FromWT}.

Models based on word embeddings have been used to track semantic changes over time. These models usually construct the embeddings separately in each time period and then align the vectors across time, e.g. as done by \citet{2016arXiv160509096H} and \citet{Kulkarni:2015:SSD:2736277.2741627}. Alternative approaches are given by \citet{2019arXiv190601688D} who use temporal referencing instead of alignment, and by \citet{Rudolph:2018:DEL:3178876.3185999} who use dynamic embeddings (based on exponential family embeddings \citep{rudolph2016exponential}) where word embeddings are set in a probabilistic framework. A different dynamic embedding model (called dynamic Skip-gram) is given by \citet{Bamler:2017:DWE:3305381.3305421} who use a Kalman filter prior to connect embeddings across time periods.

It is not straightforward to ascertain which of the three approaches, if any, is the best. Whilst word embedding models appear to be the most popular category for semantic modelling \citep{kutuzov2018diachronic}, these models tend to admit either multiple word senses or multiple time periods but not both simultaneously. To the best of our knowledge, there is currently no embedding model that allows multiple word senses to be modelled consistently across time. The dynamic embedded topic model \citep{dieng2019dynamic} tracks document topics, but not word senses, across time using word embeddings; so although it is not in substance a model for sense change, it is a step in this direction. Moreover, the embedding and graph-based models are not stochastic-process-based generative models, and this limits their interpretability and Bayesian measures of uncertainty. In contrast, the topic-based SCAN and GASC models are generative, admitting both multiple word senses and multiple time periods, and the model parameters have simple physical interpretations. The main drawback of SCAN and GASC is that they over-parameterise when the interaction between sense and time is weak or weakly evidenced by the data. They are particularly difficult to fit on sparse and noisy data, which are common, and where the parameterisation leads to ridge structures and multi-modality in the posterior. Our model, with an additive effect of sense and time, offers a lower-dimensional alternative.

Quantification of uncertainty in sense change estimates is rare in the field of semantic change detection, and indeed has not been attempted by the authors of SCAN and GASC whose work we build upon. The participating models in the recent SemEval shared-task on semantic change detection \citep{schlechtweg2020semeval} give a snapshot of current practice. Few models, if any, attempted to quantify uncertainty. It was not a SemEval assessment criterion, as is typical in this literature.

\section{Data}
\label{sec:data}

Consider the three text snippets containing the word "bank" in Table \ref{tab:snippet_examples}, where "bank" is used in the sense of a river-bank in the first example, and in the sense of a financial institution in the other two. The first two snippets, written in the time period 1990-2010, are taken from the non-fiction genre, whereas the third snippet, written in the time period 1830-1850, is taken from the magazine genre. The snippets are of equal length with 7 words on either side of "bank". The words in \textcolor{blue}{blue} are stopwords, i.e. the most common words in the language, which generally do not contribute to the meaning of the target word if we ignore syntax. The words in \textcolor{orange}{orange} on the other hand appear with a low frequency (called "hapaxes" if they appear exactly once), and are uninformative in the context of the models we consider.

\begin{table}
\caption{\label{tab:snippet_examples} Example text snippets for target word "\textcolor{red}{bank}"}
\begin{tcolorbox}
" . . . \textcolor{orange}{China}\textcolor{darkgray}{.} \textcolor{blue}{The} \textcolor{orange}{Yellow} River \textcolor{blue}{had} burst \textcolor{blue}{its} \textcolor{red}{banks}\textcolor{darkgray}{,} submerging vast \textcolor{blue}{areas of} \textcolor{orange}{farmland}\textcolor{darkgray}{,} washing \textcolor{blue}{away} . . . " \\
{\small \hspace*{\fill} --- "1421: the year China discovered America" (2003) -- non-fic -- Menzies, Gavin}\\

" . . . \textcolor{blue}{to} examine \textcolor{blue}{whether} institutions \textcolor{blue}{like the World} \textcolor{red}{Bank} \textcolor{blue}{and the} International Monetary Fund \textcolor{blue}{needed} \textcolor{orange}{restructuring} . . . " \\
{\small \hspace*{\fill} --- "The price of loyalty: George W. Bush, the White House, and the education \\ \hspace*{\fill} of Paul O'Neill" (2004) -- non-fic -- Suskind, Ron}\\

" . . .  subject \textcolor{blue}{of} \textcolor{orange}{continuing} specie payments\textcolor{darkgray}{.} \textcolor{blue}{Though the} \textcolor{red}{Bank} \textcolor{blue}{of the United States had previously} determined . . . " \\
{\small \hspace*{\fill} --- "Philadelphia Banking" (1839) -- mag -- US Democratic Review: Nov 1839}
\end{tcolorbox}
\end{table}

For a given target word, we define the data $W$ as a collection of $D$ snippets with a symmetric context window of $L/2$ words on either side of the target word (ignoring sentence and paragraph boundaries), with the stopwords, uninformative words and punctuation removed, and the words lemmatised (i.e. replaced with root words such as "wash" instead of "washing" in the first example). The snippets span multiple discrete and contiguous time periods, and may be taken from any number of text genres. Our model, described in the next section, could be applied to any dataset with these features. We analyse two real datasets in this paper: the context data for target word "bank" extracted from the Corpus of Historical American English (COHA) published by \citet{davies2010corpus}, which is used as an illustrative example, and the context data for target word "kosmos" extracted from the Diorisis Ancient Greek Corpus published by \citet{TheDiorisisAncientGreekCorpus}, which is the focus of this work. We additionally use synthetic data to compare the models' predictive performance on held-out true sense labels.

\begin{table}
\caption{\label{tab:bank_snippet_counts} Frequencies of "bank" in each sense-time block across all genres}
\centering
\begin{tabular}{l r r r r r r r r r r}
    \toprule
    \multirow{2}{*}{Sense} & \multicolumn{10}{c}{Start of 20-year period} \\
                  & 1810 & 1830 & 1850 & 1870 & 1890 & 1910 & 1930 & 1950 & 1970 & 1990\\
    \midrule
    River-bank  & 193 &  95 & 135 & 126 & 175 & 123 &  86 &  89 &  65 & 73\\
    Institution &  89 & 189 & 201 & 261 & 199 & 258 & 298 & 275 & 307 & 288\\
    \bottomrule
\end{tabular}
\end{table}

\begin{table}
\caption{\label{tab:kosmos_snippet_counts} Frequencies of "kosmos" in each sense-genre-time block}
\centering
\begin{tabular}{l l r r r r r r r r r}
    \toprule
    \multirow{2}{*}{Genre} & \multirow{2}{*}{Sense} & \multicolumn{9}{c}{Century} \\
                  & & --7 & --6 & --5 & --4 & --3 & --2 & --1 &   1 &  2\\
    \midrule
    \multirow{3}{*}{Narrative}     & Decoration & 0 & 0 &  5 & 13 & 10 &  0 & 45 & 123 & 27\\
                                   & Order      & 0 & 0 & 10 & 14 &  9 &  0 & 68 &  57 & 17\\
                                   & World      & 0 & 0 &  0 &  0 &  3 &  0 & 31 &  20 &  2\\
    \midrule
    \multirow{3}{*}{Non-narrative} & Decoration & 2 & 0 &  9 & 51 &  4 & 37 & 33 &  43 &  2\\
                                   & Order      & 1 & 0 &  1 & 29 &  6 &  1 &  2 &  25 &  3\\
                                   & World      & 0 & 0 &  0 & 52 &  3 & 26 & 15 & 303 & 35\\
    \bottomrule
\end{tabular}
\end{table}

The "bank" example was used by \citet{frermann2016bayesian}, and we use it as an illustrative example due to its relative simplicity since the two main senses of "bank" are very distinct. There are c.\,74,000 instances of "bank" in COHA covering the years 1810-2009, which we divide into ten 20-year contiguous blocks, and spanning four genres (fiction, non-fiction, news and magazine). We extract snippets of length $L=14$ words (not counting the target word) around these instances. The snippet length has to strike a balance between including meaningful nearby words and not including noise from distant words, and we found the length $L=14$ to be sufficient for a human to identify the sense in most cases. For computational ease, we randomly subsample a maximum 100 snippets per genre-time block, giving us 3,685 selected snippets and c.\,70,000 non-selected snippets. We manually tag 3,525 of the selected snippets with the correct sense of "bank", grouping together the related meanings of river-bank, edge, tilt or heap, and the related meanings of a financial institution or a store (e.g. blood bank). The remaining snippets were either ambiguous or used "bank" as a proper noun (e.g. Mr Banks), or a very small number of other senses. We identify and remove stopwords using the R package \texttt{Stopwords} \citep{stopwords2020} as well as part-of-speech tags, marking anything other than nouns, adjectives, verbs and adverbs as stopwords. We further restrict the data to the top 70\% most frequently occurring words in the non-selected snippets. This is done for efficiency reasons, and seems to incur little loss of information. We refer to the 30\% of words omitted as "uninformative" words: they occur infrequently in the target context. The observation model for context words in the selected snippets is not affected by this registration, as uninformative words are defined by their frequency in non-selected snippets. Using this registration criterion, we have a vocabulary of 973 words.

The "kosmos" example was used by \citet{GASC_2019arXiv190305587P}, and we use it as our test case because, in contrast to "bank", we found it particularly challenging to analyse using existing models and tools. "Kosmos" can be used in one of three senses, i.e. decoration, order or world, and expert sense-annotation is provided by \citet{vatri_lahteenoja_mcgillivray_2019}. We use snippets of length $L=14$ as before and, following \citet{GASC_2019arXiv190305587P} for testing purposes, retain only the "collocates", i.e. all snippet instances where an ancient Greek expert was able to identify the sense based on contextual information alone. A "real-use" analysis including non-collocates is given Appendix~\ref{sec:appendix:kosmos}. We group the text genres into "narrative" and "non-narrative", and divide time into 9 contiguous centuries from 700 BC to AD 200. We remove stopwords in the same way as for "bank" plus the additional stopwords identified by \citet{greek_stopwords}. Hapaxes (i.e. words that appear only once in the selected snippets) are removed, approximating the observation model as we do not condition on the event that a context word appears at least twice over all selected snippets. This is in contrast to the registration process for "bank" where we had a large set of non-selected snippets. This leaves us with 1,144 snippets and a vocabulary of 968 words appearing in snippets associated with "kosmos". Tables~\ref{tab:bank_snippet_counts}--\ref{tab:kosmos_snippet_counts} show that the "kosmos" data is a lot more sparse and fragmented than the "bank" data, especially for the early time periods.

\section{Prior and observation models}
\label{sec:the_model}

In this section we introduce our new DiSC model and highlight how it improves upon the existing GASC model, noting that GASC is the same as SCAN except that it allows the sense prevalence to vary according to the text genre in addition to time. DiSC, given in Algorithm~\ref{alg:DiSC_generative_model}, is a generative model of how the context words in the snippets around a given target word are emitted from a latent stochastic process. We make the simplifying assumption commonly used in NLP that each snippet is a "bag-of-words" (i.e. word order and grammar are ignored) of length $L$, not counting the target word itself. Stopwords and uninformative words are generated in any snippet $d$ with probabilities $q^\text{SW}$ and $q^\text{U}$ respectively, so the number of context words that are neither stopwords nor uninformative, denoted $L_d$, has a binomial distribution $L_d | L, q^\text{SW}, q^\text{U} \sim \Bin(L, 1 - q^\text{SW} - q^\text{U})$. The context positions occupied by these words are a random subset $\{ i_1 , \dots, i_{L_d} \}$ of size $L_d$ drawn from $\{1, \dots, L\}$, i.e. the order of words is irrelevant as per the bag-of-words assumption. Once we condition on the $L_d$ values, the likelihood (which we write down in Section~\ref{sec:inference} below) does not depend on $q^\text{SW}$ and $q^\text{U}$. We can therefore drop $q^\text{SW}$ and $q^\text{U}$ from all analysis hereafter.

The snippets span $T$ discrete and contiguous time periods and $G$ text genres, so we have deterministic mappings between each snippet $d$ and its time and genre labels $\tau_d \in \{1,\dots,T\}$ and $\gamma_d \in \{1,\dots,G\}$ respectively. The target word can be used in one of $K$ senses in any snippet $d$, so the single sense assignment $z_d$ for the whole snippet is emitted as a draw from a multinomial distribution over the senses $\{1, \dots, K\}$ parameterised by the $K$-dimensional probability vector $\tilde{\phi}^{\gamma_d,\tau_d}$, that is $z_{d} | \tilde{\phi}^{\gamma_d,\tau_d} \sim \Mult \left( \tilde{\phi}^{\gamma_d,\tau_d}_{1}, \dots, \tilde{\phi}^{\gamma_d,\tau_d}_{K} \right)$. The vocabulary consists of $V$ words so, given the sense assignment $z_d$, for each context position $i \in \{ i_1 , \dots, i_{L_d} \}$ the context word $w_{d,i}$ is emitted as a draw from a multinomial distribution over the words $\{1, \dots, V\}$ parameterised by the $V$-dimensional sense-dependent probability vector $\tilde{\psi}^{z_d,\tau_d}$, that is $w_{d,i} | z_d, \tilde{\psi}^{z_d,\tau_d} \sim \Mult \left( \tilde{\psi}^{z_d,\tau_d}_{1}, \dots, \tilde{\psi}^{z_d,\tau_d}_{V} \right)$. At this level DiSC is the same as GASC. 

In contrast, the dynamic topic model would have $z_{d,i} | \tilde{\phi}^{\gamma_d,\tau_d} \sim \Mult \left( \tilde{\phi}^{\gamma_d,\tau_d}_{1}, \dots, \tilde{\phi}^{\gamma_d,\tau_d}_{K} \right)$ independently for each context position $i$. If we modify this and impose a single topic per document, it becomes precisely the GASC model with $G=1$; so comparison of DiSC with the dynamic topic model itself is not appropriate.

The probability vectors $\tilde{\phi}^{g,t}$ and $\tilde{\psi}^{k,t}$ are softmax transforms of the real-valued sense prevalence parameter vector $\phi^{g,t}$ and word parameter vector $\psi^{k,t}$ respectively, i.e. 
\begin{equation}
\tilde{\phi}^{g,t} = \frac {\exp (\phi^{g,t})} {\sum_{k=1}^K \exp (\phi_{k}^{g,t})}
\qquad \text{and} \qquad 
\tilde{\psi}^{k,t} = \frac {\exp (\psi^{k,t})} {\sum_{v=1}^V \exp (\psi_{v}^{k,t})}.\label{eq:softmax}
\end{equation}
Here $\tilde{\phi}, \phi$ are $K \times G \times T$ dimensional arrays and $\tilde{\psi}, \psi$ are $V \times K \times T$ dimensional arrays. The latent variables $\phi$ and $\psi$ are not identifiable in this setup, but this is not an issue since these are only used to determine the priors for the probability arrays $\tilde{\phi}$ and $\tilde{\psi}$ which are the variables of interest. The word parameter vector $\psi^{k,t}$, which depends on both the sense $k$ and the time $t$, is defined as the sum of a word-sense parameter vector $\chi^k$ and a word-time parameter vector $\theta^t$. We place independent autoregressive AR(1) priors on the elements of $\phi^{g,t}$ and $\theta^t$, and independent normal priors on the elements of $\chi^k$, as defined in Algorithm \ref{alg:DiSC_generative_model} lines \ref{alg_line:priors_start}-\ref{alg_line:priors_end}. The prior hyperparameters are kept fixed. 

\begin{algorithm}[!t]
\caption{DiSC: generative model}
\label{alg:DiSC_generative_model}
\begin{algorithmic}[1]

\centering
\Statex ------------------------ PRIOR MODEL ------------------------ 

\justifying
\State fix hyperparameters $\kappa_{\phi}, \kappa_{\theta}, \kappa_{\chi}, \alpha_{\phi}, \alpha_{\theta}$ (with $|\alpha_{\phi}| < 1, |\alpha_{\theta}| < 1$) 

\State initialise at time $t=1$ \label{alg_line:priors_start}
    \For {genre $g \in 1:G$}
        \State draw sense prevalence parameter $\phi^{g,1} | \kappa_{\phi}, \alpha_{\phi} \sim \mathcal{N} \left( 0, \diag \left( \frac{\kappa_{\phi}} {1 - (\alpha_{\phi})^2} \right) \right)$
    \EndFor
    \State draw word-time parameter $\theta^{1} | \kappa_{\theta}, \alpha_{\theta} \sim \mathcal{N} \left( 0, \diag \left( \frac{\kappa_{\theta}} {1 - (\alpha_{\theta})^2} \right) \right)$

\For {time $t \in 2:T$}
    \For {genre $g \in 1:G$}
        \State draw sense prevalence parameter $\phi^{g,t} | \phi^{g,t-1}, \kappa_{\phi}, \alpha_{\phi} \sim \mathcal{N} \left( \alpha_{\phi} \phi^{g,t-1}, \diag ( \kappa_{\phi} ) \right)$
    \EndFor
    \State draw word-time parameter $\theta^{t} | \theta^{t-1}, \kappa_{\theta}, \alpha_{\theta} \sim \mathcal{N} \left( \alpha_{\theta} \theta^{t-1}, \diag ( \kappa_{\theta} ) \right)$
\EndFor

\For {sense $k \in 1:K$}
    \State draw word-sense parameter $\chi^k | \kappa_{\chi} \sim \mathcal{N} \left( 0, \diag(\kappa_{\chi}) \right)$
\EndFor \label{alg_line:priors_end}

\For {sense $k \in 1:K$ and time $t \in 1:T$}
    \State set word parameter $\psi^{k,t} = \chi^k + \theta^t$ \label{alg_line:psi}
\EndFor

\State using softmax \eqref{eq:softmax}, transform real arrays $\phi$ and $\psi$ into probability arrays $\tilde{\phi}$ and $\tilde{\psi}$

\centering
\Statex  ------------------------ OBSERVATION MODEL ------------------------ 

\justifying
\State fix probabilities of drawing stopwords $q^\text{SW}$ and uninformative words $q^\text{U}$

\For {snippet $d \in 1:D$}
    \State draw number of context words $L_d | L, q^\text{SW}, q^\text{U} \sim \Bin(L, 1 - q^\text{SW} - q^\text{U})$
    \State draw a random subset $\{ i_1 , \dots, i_{L_d} \}$ of size $L_d$ from $\{1, \dots, L\}$
    \State draw sense assignment $z_{d} | \tilde{\phi}^{\gamma_d,\tau_d} \sim \Mult \left( \tilde{\phi}^{\gamma_d,\tau_d}_{1}, \dots, \tilde{\phi}^{\gamma_d,\tau_d}_{K} \right)$
    \For {context position $i \in \{ i_1 , \dots, i_{L_d} \}$}
        \State draw context word $w_{d,i} | z_d, \tilde{\psi}^{z_d,\tau_d} \sim \Mult \left( \tilde{\psi}^{z_d,\tau_d}_{1}, \dots, \tilde{\psi}^{z_d,\tau_d}_{V} \right)$
    \EndFor
\EndFor

\end{algorithmic}
\end{algorithm}

The GASC generative model is given in Algorithm~\ref{alg:GASC_generative_model} in Appendix~\ref{sec:appendix:gasc_model}. Our DiSC model differs from GASC in three main ways. The fundamental difference is that we assume the effects due to sense and time are additive: the 3-dimensional $V \times K \times T$ array $\psi$ in GASC is replaced by two 2-dimensional arrays in DiSC, i.e. a $V \times K$ array $\chi$ and a $V \times T$ array $\theta$, so that we have $\psi^{k,t} = \chi^k + \theta^t$ for $k \in \{1,\dots,K\}$ and $t \in \{1,\dots,T\}$. This reduces the dimension of $\psi$ from $VKT$ parameters in GASC to $V(K+T)$ parameters in DiSC. The implications of this for sense change measurements are discussed in Section~\ref{sec:sense_change}.

The second difference is in the modelling of the time series variables $\phi^{g,t}, \theta^t$ in DiSC and $\phi^{g,t}, \psi^{k,t}$ in GASC. Our priors on the time series are autoregressive AR(1) processes with proper stationary distributions, whereas GASC uses improper priors without global means or stationary distributions. For example, the prior distribution of $\phi^{g,t}_k$ in GASC is $\phi^{g,t}_k | \phi^{g,t-1}_k, \kappa_{\phi} \sim \mathcal{N} \left( \phi^{g,t-1}_k, 2\kappa_{\phi} \right)$ for $t \in \{2,\dots,T\}$, and an improper uniform distribution over all real numbers for $t=1$. It is thus possible for the posterior $\phi^{g,t}_k | W$ to drift off to $\pm \infty$ since there is no global mean to tether the distribution. In contrast, the DiSC priors have a stationary distribution $\phi^{g,t}_k | \kappa_{\phi}, \alpha_{\phi} \sim \mathcal{N} \left( 0, \frac{\kappa_{\phi}} {1 - (\alpha_{\phi})^2} \right)$ for all $t$. We expect this to lead to more homogeneous behaviour at the beginning and end of the time series.

The third difference is that whilst we treat the prior hyperparameter $\kappa_\phi$ as fixed, GASC uses a random hyperprior $\kappa_{\phi} \sim \InvGamma(a,b)$ with $a,b$ fixed. However, the data do not inform this parameter at all well: $\kappa_\phi$ is conditionally independent of the data $W$ given $\phi$, we have a relatively small number of $\phi$ parameters, and these are in turn conditionally independent of the data $W$ given the unknown sense labels $z$. In contrast, physical considerations lead to an informative prior for $\kappa_\phi$. The joint posterior \eqref{eq:posterior_line1} below is insensitive to the choice of $\kappa_\phi$ for any plausible value of this parameter, so we can fix $\kappa_\phi$ without loss using the prior elicitation described in Section~\ref{sec:hyperparameter_settings}.

\subsection{Sense representation and sense change}
\label{sec:sense_change}

Following the framework of \citet{frermann2016bayesian} for these bag-of-words models, we define sense $k$ of the target word at time $t$ as the distribution $\big( \tilde{\psi}^{k,t}_1, \dots, \tilde{\psi}^{k,t}_V \big)$ over words $\{1,\dots,V\}$ appearing in the context of the target word. "Sense change" may therefore be defined as the evolution of the $V \times K$ matrix $\tilde{\psi}^{\cdot,t}$ over $t \in \{1,\dots,T\}$. Similarly, "sense prevalence change" may be defined as the evolution of the $K \times G$ matrix $\tilde{\phi}^{\cdot,t}$ over $t \in \{1,\dots,T\}$. We loosely refer to both evolutions as "sense change" for brevity. This definition of sense change does not correspond to any sort of real meaning change (which we do not define, but we have in mind something like changes in the dictionary definition of the target word), which is a limitation of all bag-of-words models. 

We parameterise DiSC to model key structural drivers of sense difference (over senses) and sense change (over time). The additive main effects in DiSC ($\chi$ and $\theta$ respectively) correspond to these drivers. This allows us to model the main effects, and the data to inform them, in a direct way. In contrast, GASC models the effects and their interaction in a general way without an \textit{explicit} parameterisation or modelling of the main effects. Useful structural information is left out of the prior, as the model imposes no core additive structure. We describe below the main data features that we see as informing sense difference and sense change.

The probability $\tilde{\psi}^{k,t}_v$ of context word $v\in\{1,\dots,V\}$ being used in a snippet can change when the background usage frequency of word $v$ changes in the text corpus taken as a whole. For example, the background usage frequency of the word "telephone" likely increases over the 20\textsuperscript{th} century. It is then more likely to appear in any context, across all target word senses. This is, in our definition, a form of sense change. We expect this time-effect, captured by $\theta^{t}$ in DiSC, to be a common driver of sense change in these models. The time-effect is not explicitly modelled in GASC.

Similarly, we expect certain words to appear more frequently in the context of particular senses of the target word regardless of their background usage frequency. Examples include "water" and "money" for the two senses of "bank" respectively. This relatively strong sense-effect informs sense difference for the target word, and is basic to the bag-of-words setup. DiSC, in contrast to GASC, models this sense-effect explicitly through $\chi^k$.

On the other hand, we might expect certain words to change in frequency at different rates in the context of different senses of the target word. Such changes might be driven by actual changes in the meaning of the target word. For example, "telephone" might increase in frequency more rapidly in the financial institution sense of "bank", as the sense changes to reflect more modern ways of banking, than it does in the river-bank sense. This sense-time interaction effect is captured by GASC but not by DiSC. However, GASC does not distinguish between the main and interaction effects, so if we are interested in isolating this behaviour then it is necessary to remove the main effects. We illustrate this in our analysis of the "bank" data in Section~\ref{sec:experiment_bank} below. 

Goodness-of-fit is generally retained in DiSC even in the presence of a real interaction. Words contributing to poor fit due to missed interactions are words that \textit{both} appear frequently in the context of multiple target word senses \textit{and} evolve in frequency at different rates, since these have both large percentage change and large value. Large percentage errors in context probability for very small context probabilities are of little consequence. This feature of the data makes DiSC relatively robust to interaction in practice, as demonstrated in the synthetic data experiments in Section~\ref{sec:experiment_synthetic} below. 

The additive structure in DiSC helpfully prevents switching of sense labels between time periods. In MCMC targeting GASC, the senses can settle into different sense-label permutations in different time periods, as $\psi$-values are only loosely connected temporally via their priors. This is a particular problem when some context words appear frequently across more than one sense. In such cases, the MCMC may get stuck in a local mode where the sense labels are not aligned across time. This is no criticism of GASC, but adds to the challenge of fitting it to data.

One basic and important kind of sense change, i.e. the emergence of a new sense in addition to the existing senses, is captured in both models if $K$ is large enough to accommodate the new sense. The new sense $k$ exists at all times, but its prevalence $\tilde{\phi}^{g,t}_k$ changes from being very small to significant as it emerges with increasing $t$.

\subsection{Hyperparameter settings}
\label{sec:hyperparameter_settings}

We choose the number of senses $K$ equal to the smallest value such that each sense $k \in \{1,\dots,K\}$ in the model output, as identified by the most probable words under that sense, is distinct and meaningful in the judgement of the expert user. This is practical when DiSC is used as an exploratory tool to discover the lifespan and usage rate of distinct senses without the need for hand-labelling. Alternatively, the expert user may have hand-labelled a small set of snippets, in which case we would have prior knowledge of $K$. Setting $K$ therefore requires a few trial runs. In our case, $K = 2$ for "bank" and $K = 3 \text{ or } 4$ for "kosmos" depending on the task (see Section~\ref{sec:experiment_kosmos} and Appendix~\ref{sec:appendix:kosmos}). If the value of $K$ is still in question, model selection tools may be used, for example classic Bayesian tools such as Bayes factors \citep{doi:10.1080/01621459.1995.10476572} or reversible jump MCMC \citep{Green95reversiblejump}, or rather the state-of-the-art on these such as \citet{2021arXiv210607462X} or \citet{doi:10.1080/10618600.2013.805651} respectively. However, we have not investigated this.

Implementing SCAN/GASC requires setting the hyperparameter $\kappa_\psi$ in $\psi^{k,t}_v | \psi^{k,-t}_v \sim \mathcal{N} \left( \frac{1}{2} (\psi^{k,t-1}_v + \psi^{k,t+1}_v), \kappa_\psi \right)$ and the hyperparameters $a,b$ in $\kappa_{\phi} \sim \InvGamma(a,b)$. \citet{frermann2016bayesian} used the setting $\kappa_\psi=0.1,\, a=7$ and $b=3$ for SCAN whereas \citet{GASC_2019arXiv190305587P} used $\kappa_\psi=0.01,\, a=1$ and  $b=1$ for GASC. We report results for the SCAN choice as these give the best performance for these models.

For the AR(1) processes in DiSC, we use the parameters $\alpha_\phi = \alpha_\theta = 0.9$ --- a high value --- in order to have weak mean reversion, so that we have a proper process prior without unduly influencing the posteriors. 

To set $\kappa_\phi$ in DiSC, we elicit a prior by defining what we consider to be an extreme sense prevalence difference. The number of senses $K$ is relatively small (compared to the vocabulary size $V$). Taking two fixed senses $l,m \in \{1,\dots,K\}$, we allow a difference as large as $\tilde{\phi}^{g,t}_l / \tilde{\phi}^{g,t}_m \approx 100$ to be possible but extreme. This can easily be adjusted depending on the data-modelling context. On the logit-scale, we therefore assert that $\phi^{g,t}_l - \phi^{g,t}_m > \log 100$ is a 3-sigma event. From the prior stationary distribution in the AR(1) process, $\mathbb{V} ( \phi^{g,t}_l - \phi^{g,t}_m ) = \frac{2\kappa_\phi}{1-(\alpha_\phi)^2}$; so we express our preference with $3 \left( \frac{2\kappa_\phi}{1-(\alpha_\phi)^2} \right)^\frac{1}{2} = \log 100$, giving $\kappa_\phi = 0.25$ on rounding. There is no simple comparison with $\kappa_\phi$ in SCAN/GASC due to the AR(1) structure in DiSC. If we replace the threshold for a rare event with $\tilde{\phi}^{g,t}_l / \tilde{\phi}^{g,t}_m \approx 10^6$ then we find $\kappa_\phi \approx 2$: the standard deviation changes over a small range from $0.5$ to $\sqrt{2}$. Larger values are hard to justify.

The vocabulary size $V$ is typically quite large (c.~1,000 in our examples), so we might expect greater variation in the probabilities for context words in a given sense, perhaps by a factor of of 1,000. That is, for any fixed time $t$, sense $k$ and pair of words $x,y \in \{1,\dots,V\}$, the ratio of context word probabilities might be as large as $\tilde{\psi}^{k,t}_x / \tilde{\psi}^{k,t}_y \approx 1\,000$. Now $\mathbb{V} ( \psi^{k,t}_x - \psi^{k,t}_y ) = \mathbb{V} ( \chi^k_x - \chi^k_y + \theta^t_x - \theta^t_y ) = 2\kappa_\chi + \frac{2\kappa_\theta}{1-(\alpha_\theta)^2}$, so we express our preference with $3 \left( 2\kappa_\chi + \frac{2\kappa_\theta}{1-(\alpha_\theta)^2} \right)^\frac{1}{2} = \log 1\,000$. Attributing the variance in equal parts to $\chi$ and $\theta$ by setting $\kappa_\chi = \frac{\kappa_\theta}{1-(\alpha_\theta)^2}$, since they are additive effects on the same scale, we have $\kappa_\chi = 1.25$ and $\kappa_\theta = 0.25$ on rounding. As for $\kappa_\phi$, this is robust to the choice of threshold due to the logarithm, so the posterior is relatively insensitive over values plausible a priori. Moreover, we have considered two fixed context words, not the most extreme pair, so more extreme variation is allowed.

\section{Posterior distribution and MCMC inference}
\label{sec:inference}

For convenience, we infer posterior distributions for the sense prevalence parameters $\phi$, the word-time parameters $\theta$ and the word-sense parameters $\chi$, but our interest is in the identifiable probability arrays $\tilde{\phi}$ and $\tilde{\psi}$ which are given as deterministic functions of these latent variables. We expect the sense assignment vector $z$ to be of less interest, except in testing where we have hand-annotated meanings. The joint posterior for $\phi, \theta, \chi, z$ given the data $W$ is defined by 
\begin{align}
    \pi(\phi, \theta, \chi, z | W) &\propto \pi(\phi) \pi(\theta) \pi(\chi) \pi(z | \phi) p(W | z, \theta, \chi) \label{eq:posterior_line1} \\
    &= \pi(\phi) \pi(\theta) \pi(\chi) \prod_{d=1}^D \tilde{\phi}_{z_d}^{\gamma_d,\tau_d} \prod_{i=i_1}^{i_{L_d}} \tilde{\psi}_{w_{d,i}}^{z_d,\tau_d} \label{eq:posterior_line2} \\
    &= \pi(\phi) \pi(\theta) \pi(\chi) \prod_{t=1}^T \prod_{k=1}^K \Bigg( \prod_{g=1}^G ( \tilde{\phi}_k^{g,t} )^{N_{k,g,t}^{z}} \Bigg) \Bigg( \prod_{v=1}^V ( \tilde{\psi}_v^{k,t} )^{N_{v,k,t}^{W,z}} \Bigg) \label{eq:posterior_line3}
\end{align}
where $N_{k,g,t}^{z} = \sum_{d: \tau_d=t \atop \text{and } \gamma_d = g} \mathbb{I}(z_d = k)$ is the number of snippets with sense assignment $k$ and genre $g$ at time $t$, and $N_{v,k,t}^{W,z} = \sum_{d: \tau_d=t \atop \text{and } z_d = k} \sum_{i=i_1}^{i_{L_d}} \mathbb{I}(w_{d,i} = v)$ is the number of occurrences of context word $v$ across all snippets with sense assignment $k$ at time $t$. The conditional posterior for $z$ is defined, independently for each snippet $W_d$, by
\begin{equation} \label{eq:z_posterior}
    \pi(z_d | W_d, \phi, \psi) \propto \tilde{\phi}^{\gamma_d,\tau_d}_{z_d} \prod_{i=i_1}^{i_{L_d}} \tilde{\psi}^{z_d,\tau_d}_{w_{d,i}},
\end{equation}
which is a multinomial distribution over possible senses $\{1,\dots,K\}$. 

The authors of SCAN and GASC both use a blocked Gibbs strategy to alternately sample $z | W, \phi, \psi$ using \eqref{eq:z_posterior} and $\phi | z$ and $\psi | z,W$. Under the SCAN and GASC models, each column of $\tilde{\phi}$ and $\tilde{\psi}$ has a logistic normal distribution which the authors target with a Gibbs sampler based on the auxiliary uniform variable method of \citet{mimno2008gibbs}. An alternative Gibbs sampler is based on auxiliary Polya-Gamma variables \citep{PolyaGamma_2012arXiv1205.0310P} and an approximate method based on the same is given by \citet{chen2013scalable}. We describe these methods in Appendix \ref{sec:appendix:existing_methods}.

Under our DiSC model, since $\tilde{\psi}^{k,t}$ does not have a logistic normal distribution, the latter two methods cannot be used in a straightforward manner, although the auxiliary uniform variable method can be easily adapted. Moreover, both the auxiliary uniform and Polya-Gamma variable methods are very inefficient for these models, whereas the approximate method is obviously not asymptotically exact. We describe our MCMC sampler, which is asymptotically exact and at least as efficient as the existing samplers, and may be used with any of these models. 

We marginalise over the discrete $z$, removing the need to sample $z | W, \phi, \psi$, and alternate between sampling $\phi | W, \theta, \chi$ and $\theta | W, \phi, \chi$ and $\chi | W, \phi, \theta$ (or, in the case of SCAN/GASC, between $\phi | W, \psi$ and $\psi | W, \phi$). The marginal likelihood for $\phi, \theta, \chi |W$ is
\begin{align}
    p(W|\phi, \theta, \chi) &= \prod_{d=1}^D p(W_d | \phi, \theta, \chi) \label{eq:likelihood_line1}\\
    &= \prod_{d=1}^D \sum_{k=1}^K p(z_d=k | \phi) p(W_d | z_d=k, \theta, \chi) \label{eq:likelihood_line2}\\
    &= \prod_{d=1}^D \sum_{k=1}^K p(z_d=k | \phi) \prod_{i=i_1}^{i_{L_d}} p(w_{d,i} | z_d=k, \psi) \label{eq:likelihood_line3}\\
    &= \prod_{d=1}^D \sum_{k=1}^K \tilde{\phi}_k^{\gamma_d,\tau_d} \prod_{i=i_1}^{i_{L_d}} \tilde{\psi}_{w_{d,i}}^{k,\tau_d} \label{eq:likelihood_line4}
\end{align}
where \eqref{eq:likelihood_line1} exploits the conditional independence between the snippets, \eqref{eq:likelihood_line2} comes from conditioning on (and summing over) the sense assignment $z_d$ for snippet $d$, \eqref{eq:likelihood_line3} exploits the conditional independence of the context words $w_{d,i}, i \in \{i_1,\dots,i_{L_d}\}$ in snippet $d$, and \eqref{eq:likelihood_line4} simply picks up the appropriate probabilities from the $\tilde{\phi}$ and $\tilde{\psi}$ arrays. The conditional posteriors for $\phi$, $\theta$ and $\chi$ are therefore
\begin{align} 
    \pi(\phi | W,\theta,\chi) &\propto \pi(\phi) \prod_{d=1}^D \sum_{k=1}^K \tilde{\phi}_k^{\gamma_d,\tau_d} \prod_{i=i_1}^{i_{L_d}} \tilde{\psi}_{w_{d,i}}^{k,\tau_d} \text{,} \label{eq:phi_posterior}\\
    \pi(\theta | W,\phi,\chi) &\propto \pi(\theta) \prod_{d=1}^D \sum_{k=1}^K \tilde{\phi}_k^{\gamma_d,\tau_d} \prod_{i=i_1}^{i_{L_d}} \tilde{\psi}_{w_{d,i}}^{k,\tau_d} \label{eq:theta_posterior}\\  \text{and } 
    \pi(\chi | W,\phi,\theta) &\propto \pi(\chi) \prod_{d=1}^D \sum_{k=1}^K \tilde{\phi}_k^{\gamma_d,\tau_d} \prod_{i=i_1}^{i_{L_d}} \tilde{\psi}_{w_{d,i}}^{k,\tau_d} \label{eq:chi_posterior}
\end{align}
respectively, which we can sample efficiently using gradient-based MCMC methods such as Metropolis-Adjusted Langevin Algorithm (MALA) and Hamiltonion Monte Carlo (HMC). We describe these methods, including the derivation of the gradient vectors and automatic parameter tuning, in Appendix~\ref{sec:appendix:new_method}. At this level, the marginal posterior distribution in GASC is the same as DiSC (the essential difference being the parameterisation of $\psi$ and the priors) and any sampler relevant for DiSC also applies for GASC. This marginal is often tractable for LDA but not used, as the collapsed Gibbs sampler obtained by integrating over the conjugate priors of the continuous parameters is favoured there. That is not possible here as the priors are not conjugate. In addition to these sampling methods, we implemented a simple random-walk Metropolis sampler, both jointly and marginally over $z$. This was useful for code-checking not competitive and has been omitted.

For both datasets, we find that 20k MCMC iterations for SCAN/GASC with burn-in of 10k, and 10k MCMC iterations for DiSC with burn-in of 5k, are sufficient for convergence, where an iteration is one update over all parameters (except $\kappa_\phi$ for SCAN/GASC, which is updated every 50 iterations). Work to date on SCAN and GASC appears to take just 1k MCMC iterations in total, which we found was not nearly enough for convergence on these data. All figures and results reported in the following sections are based on posterior means unless otherwise indicated. The posterior is invariant under sense relabelling in all models, but this behaviour was not observed in any converging run.

\section{Experiment 1: Finding the best sampler}
\label{sec:experiment_samplers}

We take as our test case for comparing the samplers the problem of fitting the SCAN model (i.e. GASC with one genre) for the target word "bank" in COHA. We are limited in these choices since, as we report in Section~\ref{sec:experiment_kosmos}, no MCMC sampler we tried converged for the SCAN/GASC model on the "kosmos" analysis, and the Poly-Gamma samplers cannot be used with DiSC due to the additive form for $\psi$.
The metric we use for this comparison is the effective sample size (ESS) per hour of CPU time after the burn-in. We implemented the samplers in the R programming language, as efficiently as we could, and using the same functions as far as possible. We checked that they converged to the same posterior distributions for all variables with high precision on small synthetic datasets. All runs are done sequentially on the same Linux PC.

For $\tilde{\phi}$, we are inferring $KGT$ parameters and we take the median ESS per hour over all parameters. For $\tilde{\psi}$, we are inferring $VKT$ parameters but our interest is mainly in the most representative words for each sense; so we take the median ESS per hour across the $\tilde{\psi}$ parameters for the 20 most probable words under each sense marginally over all time periods. The medians, together with the interquartile ranges, are shown in Table \ref{tab:ess}. Even allowing for uneven coding efficiency, it is clear that the differences are substantial. 

\begin{table}
\caption{\label{tab:ess}Median (interquartile range) ESS per hour of CPU time}
\begin{threeparttable}
\centering
\begin{tabular}{ l r r@{}r@{}c@{}r r r@{}r@{}c@{}r }
    \toprule
    Sampling method & \multicolumn{5}{c}{ESS for $\tilde{\phi}$} & \multicolumn{5}{c}{ESS for $\tilde{\psi}$} \\
    \midrule
    Aux uniform variable        &    40 &    (&37\phantom{-} &--\phantom{-}&    44) &   419 & (&255\phantom{-} &--\phantom{-}&   604) \\
    Aux Polya-Gamma             &    81 &    (&34\phantom{-} &--\phantom{-}&   164) &    80 &  (&63\phantom{-} &--\phantom{-}&   102) \\
    Aux Polya-Gamma (approx)    & 1,557 &   (&723\phantom{-} &--\phantom{-}& 1,694) &   671 & (&546\phantom{-} &--\phantom{-}&   927) \\
    MALA                        & 1,477 & (&1,236\phantom{-} &--\phantom{-}& 1,683) &   382 & (&270\phantom{-} &--\phantom{-}&   676) \\
    HMC (5 leapfrog steps)      &   672 &   (&587\phantom{-} &--\phantom{-}&   993) & 1,105 & (&762\phantom{-} &--\phantom{-}& 1,651) \\
    MALA+HMC (variable\textsuperscript{†} steps) & 1,613 & (&864\phantom{-} &--\phantom{-}& 1,816) & 1,312 & (&1,054\phantom{-} &--\phantom{-}& 1,531) \\
    \bottomrule
    \end{tabular}
    \footnotesize{\textsuperscript{†} Randomly chooses 1 or 2 leapfrog steps for $\phi$ proposals and 1 or 5 leapfrog steps for $\psi$ proposals at each update}
\end{threeparttable}
\end{table}

The auxiliary uniform variable method (top row) used in the past to fit SCAN and GASC is the least efficient for $\tilde{\phi}$ by an order of magnitude or more, and the auxiliary Polya-Gamma variable method (second row) is by far the least efficient for $\tilde{\psi}$. The approximate Polya-Gamma auxiliary variable method (third row) is efficient, but not asymptotically exact, though fairly accurate in our experiments comparing against asymptotically exact samplers. Our MALA and HMC samplers targeting marginal posteriors are asymptotically exact methods of similar or better efficiency for $\tilde{\phi}$ and $\tilde{\psi}$ respectively, and are therefore preferred. MALA is comparable to a 1-step HMC sampler, so we may combine the strengths of MALA and HMC by randomly choosing either a 1-step or a multiple-step proposal in our HMC sampler at each update. This mixed MALA-HMC sampler (last row) is clearly the most efficient for both $\tilde{\phi}$ and $\tilde{\psi}$.

HMC, whether straight or mixed with MALA, has one additional parameter to tune, i.e. the number of leapfrog steps. Users may therefore prefer MALA for convenience, or mix MALA with the No-U-Turn HMC sampler of \citet{hoffman2014no} to avoid tuning the number of steps.
Trace plots in Figure~\ref{fig:trace_plots} in Appendix~\ref{sec:appendix:new_method} illustrate the differences in mixing rates between the samplers for a prevalence parameter (where the difference is most visible) in runs of equal time.

\section{Experiment 2: Analysis of sense change for "bank"}
\label{sec:experiment_bank}

We now measure the time-evolution of the prevalence and context word composition of the different senses of a target word in a simple example. Our objective is achieved if the posteriors for $\tilde{\psi}^{k,t}$ for $k \in \{1,\dots,K\}$ can be interpreted by a human as $K$ unique target word senses, since these senses are automatically identified over time $t \in \{1,\dots,T\}$ in the snippets. In the case of "bank", both DiSC and SCAN/GASC achieve this since the most probable words under an ordering based on the posterior expectation of time-averaged word probabilities $\frac{1}{T} \sum_{t=1}^T \tilde{\psi}^{k,t}$ from both DiSC and SCAN/GASC display the senses of river-bank and institution bank respectively:
\begin{Verbatim}[samepage=true]
k=1: river stream   water stand bank    leave  tree    creek  time reach 
k=2: bank  national note  money deposit saving reserve credit loan issue
\end{Verbatim}
The most probable words in each time period can easily be extracted if their evolution is of interest.

In order to assess the performance of the models, we consider their ability to recover the true sense labels $o_d, d \in \{1,\dots,D\}$ for the $D=3\,525$ snippets that we manually tagged (cf. Section~\ref{sec:data}). Let $\hat{p}(z_d=k)$ be the estimated value of $\mathbb{E}_{\phi,\psi|W} \big( p(z_d=k | W_d, \phi, \psi) \big)$ computed on the MCMC output.
The Brier score, in our case defined by
\begin{equation} \label{eq:brier_score}
    BS = \frac{1}{D} \sum_{d=1}^D \sum_{k=1}^K \Big( \hat{p}(z_d=k) - \mathbb{I} (o_d=k) \Big)^2 \text{,}
\end{equation}
is a proper scoring rule for multi-category probabilistic predictions $\hat{p}(z_d=k)$, ranging from 0 (best) to 2 (worst), which we use as a criterion for model comparison. If we set $\hat{p}(z_d=k) = \frac{1}{K}$ for all $d,k$, we get $BS = \left(1-\frac{1}{K}\right)^2 + (K-1)\left(\frac{1}{K}\right)^2 = 0.5$ in the case of $K=2$ senses for "bank", so a model must produce a score lower than this in order to be useful. Using the posterior mean probabilities $\hat{p}(z_d=k)$ obtained from normalising \eqref{eq:z_posterior}, we get the Brier scores shown in Table \ref{tab:brier_scores} from running DiSC and GASC under various genre configurations, where we fit both models using MALA. Both models therefore identify the true sense with a good level of accuracy. However, DiSC has slightly better performance in all cases.

\begin{table}
\caption{\label{tab:brier_scores} Brier scores under different genre settings}
\centering
\begin{tabular}{l l r r}
    \toprule
    $G$ & Genre grouping & DiSC & GASC \\
    \midrule
    1 & all combined & 0.152 & 0.183 \\
    2 & fiction vs others & 0.183 & 0.195 \\
    2 & fic \& non-fic vs news \& mag & 0.153 & 0.166 \\
    4 & fic vs non-fic vs news vs mag & 0.179 & 0.182 \\
    \bottomrule
\end{tabular}
\end{table}

We also examine confusion matrices from a simple classification task. We assign snippet $d$ the sense $l = \argmax_{k \in 1:K} \hat{p}(z_d = k)$. Table \ref{tab:bank_confusion_matrix} shows the key statistics from the confusion matrices produced using this decision rule (treating river-bank as the positive and institution bank as the negative condition) suggesting that the DiSC model is marginally better than SCAN despite having about half as many parameters.

\begin{table}
\caption{\label{tab:bank_confusion_matrix} "Bank" confusion matrix statistics}
\centering
\begin{tabular}{l r r r}
    \toprule
        & Sensitivity (true $+$ve & Specificity (true $+$ve & \multirow{2}{*}{Accuracy} \\
        & for river-bank sense) & for institution sense) & \\
    \midrule
    DiSC & 0.935 & 0.877 & 0.896 \\
    SCAN & 0.926 & 0.874 & 0.891 \\
    \bottomrule
\end{tabular}
\end{table}

Marginal Highest Posterior Density (HPD) intervals are a useful visualisation of uncertainty in the model output. In order to form a summary of the evolution of the sense prevalence, we look at the 95\% HPD intervals for the marginal posteriors $\tilde{\phi}^{g,t}_k | W$ from the DiSC and SCAN/GASC output. We compare these against independent well-informed estimates. We have the true sense labels $o_d, d \in \{1,\dots,D\}$ (not available in general) for these data. However, these are only multinomial draws with category-probabilities given by the unknown true prevalence $\tilde \Phi$ say. Previous authors have benchmarked against empirical sense probabilities $N_{k,g,t}^{o} / \sum_{l=1}^K N_{l,g,t}^{o}$ computed using true sense labels. These are MLEs for the components of $\tilde \Phi$ in each sense, genre and time period. We expect our HPDs, computed on {\it unlabelled} data, to match these when the counts are high and the multinomial MLEs have small errors. However, when the labelled data counts are low (as in the "kosmos" data) or vary over a wide range from one time period to another, we should quantify the uncertainty in these independent estimates of the ground-truth prevalence $\tilde \Phi$.

\begin{figure}[!t]
\includegraphics[width = 1\textwidth, keepaspectratio]{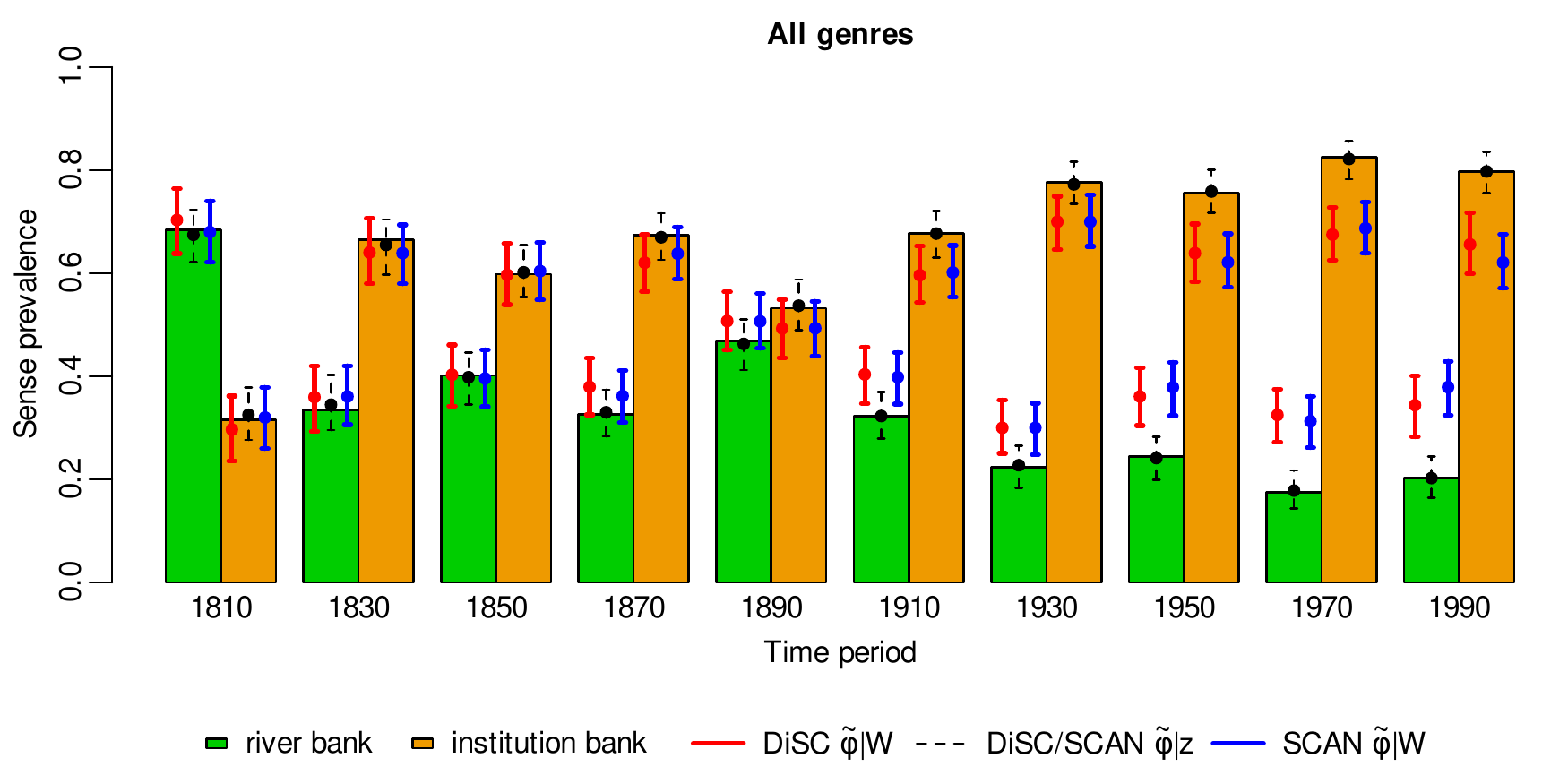}
\vspace*{-20pt}
\caption{"Bank" expert-annotated empirical sense prevalence (coloured bars with height $N_{k,g,t}^{o} / \sum_{l=1}^K N_{l,g,t}^{o}$ for each $k,g,t$) with 95\% HPD intervals (error bars) and posterior means (circles) from the model output. Note that there is no perceptible difference between the posteriors $\tilde{\phi}|z$ from DiSC and SCAN.}
\label{fig:bank_error_bars}
\end{figure}

We therefore treat the true sense labels as a second dataset, and validate our HPD sets estimated on the unlabelled data against HPD intervals for $\tilde{\phi}^{g,t}_k | (z=o)$ computed by taking the true sense labels as data in DiSC. (Whilst this may appear to favour DiSC, in fact, for this part of the model, the conditional distributions of $\tilde{\phi}|z$ are the same in all important respects, differing only by the AR(1) smoothing in DiSC.) A high degree of overlap between the HPD intervals from the two posteriors (based on labelled and unlabelled data) indicates good model performance. Figure~\ref{fig:bank_error_bars} shows that this is the case for both DiSC and SCAN (recall that SCAN is just GASC for $G=1$) for both senses of "bank" for most time periods, although they start to diverge towards the end, perhaps indicating some over-smoothing across time. The performance of DiSC and SCAN is very similar. The ground-truth prevalence estimates based on MLEs (coloured bars) and labelled-data posteriors (dashed error bars) are very close for these data as the sample sizes are large (cf. Table~\ref{tab:bank_snippet_counts}). Equivalent graphs for thinner time intervals are given in Appendix \ref{sec:appendix:bank}.

\begin{figure}[!t]
\includegraphics[width = 1\textwidth, keepaspectratio]{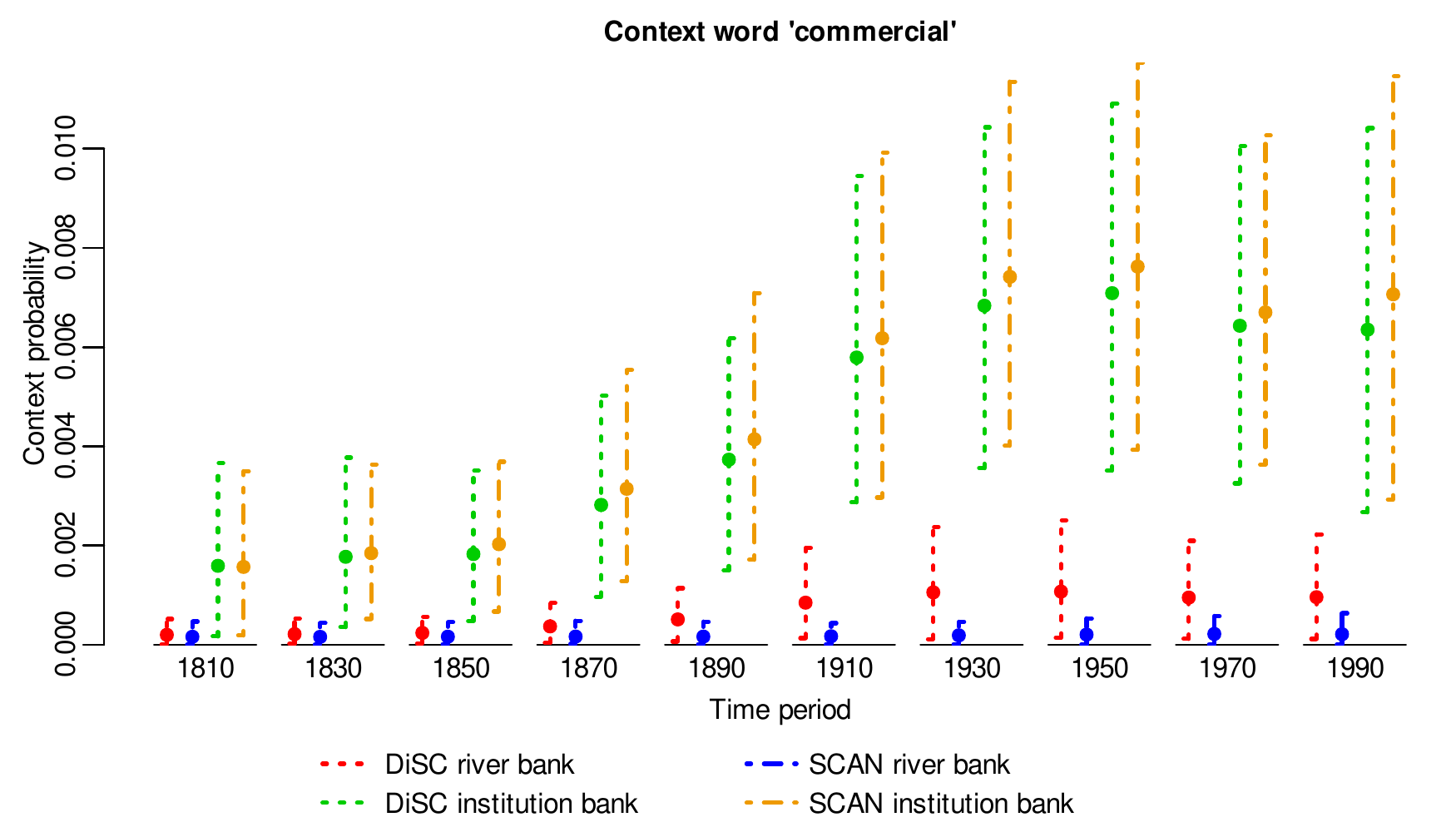}
\vspace*{-20pt}
\caption{95\% HPD intervals (error bars) and posterior means (circles) for $\tilde{\psi}_v^{k,t} | W$}
\label{fig:bank_commercial_error_bars}
\end{figure}

As discussed in Section~\ref{sec:sense_change}, DiSC captures only the main sense and time effects via the additive structure in $\psi$ whereas SCAN includes the sense-time interaction effect. As an example, Figure~\ref{fig:bank_commercial_error_bars} shows the HPD intervals for $\tilde{\psi}_v^{k,t} | W$ for the context word $v=$~"commercial" under both models. "Commercial" appears predominantly under the institution sense of "bank", and increases in probability over time, which is reflected under both posteriors. However, whilst $\tilde{\psi}_v^{k,t} | W$ under SCAN for the river-bank sense remains relatively flat, in contrast, the DiSC posterior has $\tilde{\psi}_v^{k,t} | W$ increasing for both senses, as the contribution from $\theta_v^t|W$ increases and does not distinguish sense. We know from Brier scores and confusion matrices that DiSC gives better automatic sense-labelling than SCAN or GASC on these criteria. The reason for this is that although SCAN is capturing the interaction here, and DiSC is not, such interactions seem to be rare, and when present involve a context word with a relatively high context frequency $\tilde{\psi}_v^{k,t} | W$ in the evolving sense and low context frequencies in all other senses (as generic context words have low frequency in any given sense). Our example illustrates this. Small absolute errors in small, uninformative context frequencies are offset by more reliable estimation of larger, more informative context frequencies. This is discussed further in the next section.

\section{Experiment 3: Model predictive performance on synthetic data}
\label{sec:experiment_synthetic}

Since DiSC drops sense-time interaction (cf. Sections~\ref{sec:sense_change}), it is of interest to find examples in which this causes it to fail. We therefore compare the models' predictive performance on held-out sense labels in the presence of a \textit{known} sense-time interaction effect in the data. We try to make this easy for SCAN and hard for DiSC: we calculate Brier scores on synthetic datasets of varying sizes generated using the SCAN model. Using the settings $T=9$, $K=3$ and $V \approx 1\,000$ seen in "kosmos", we sample parameters $\phi,\psi,\kappa_\phi$ according to the SCAN prior model, as this generates sense-time interaction. We then sample the true sense labels $o$ and snippets $W$ using the SCAN/DiSC observation model for a fixed number of snippets $D/T$ per time period, using the same stopword probability and context-word registration criterion as for "kosmos" (cf. Section~\ref{sec:data}). We choose a large $D/T$ so that the data contain lots of sense-time interaction and also strongly inform this interaction. This setup might be expected to favour SCAN and challenge DiSC.

These random synthetic datasets do contain an abundance of sense-time interaction. Consider the following simple measure of the level $\Lambda$ of interaction in the simulated data. First, we calculate the empirical probabilities $\hat{\tilde{\psi}}_v^{k,t} =  \frac{N_{v,k,t}^{W,o}}{N_{\cdot,k,t}^{W,o}}$ for all $v,k,t$ where $N_{\cdot,k,t}^{W,z} = \sum_{v=1}^V N_{v,k,t}^{W,z}$. Then, for each context word $v \in \{1,\dots,V\}$, we fit a linear model
\begin{equation} \label{eq:psi_linear_model}
    \hat{\tilde{\psi}}_v^{k,t} = \alpha_v + \beta_v t + \sum_{l=2}^K \delta_v^l \mathbb{I}(l=k) + t \sum_{l=2}^K \eta_v^l \mathbb{I}(l=k) + \varepsilon_v
\end{equation}
where time $t$ is continuous (since discrete $t$ leads to a perfect fit) and $\varepsilon_v$ is Gaussian noise. We make an \textit{F}-test for the interaction effects $\eta_v^l$ at level 5\% and measure the extent of sense-time interaction $\Lambda$ as the proportion of context words $\{1,\dots,V\}$ for which the interaction effects in \eqref{eq:psi_linear_model} are significant. This is a conservative measure, since it only captures linear sense-time interactions and misses potential situations where context probability increases and then decreases (or vice versa) over time. In the synthetic data with $D/T$ equal 100 and 500, the interaction level is $\Lambda = 0.13$ and $\Lambda = 0.20$ respectively. For comparison, using this measure, we have $\Lambda = 0.144$ and $\Lambda = 0.070$ respectively for the "bank" and "kosmos" datasets, so on this simple measure the extent of interaction in these synthetic data is representative. 

\begin{table}
\caption{\label{tab:synthetic}Model performance for DiSC and SCAN on synthetic data}
\begin{threeparttable}
\centering
\begin{tabular}{ l c r c r }
    \toprule
     & \multicolumn{2}{c}{$D/T = 100, \Lambda=0.131$} & \multicolumn{2}{c}{$D/T = 500, \Lambda=0.204$} \\
     & Converges? & Brier score & Converges? & Brier score \\
    \midrule
    DiSC & Yes & 0.017 & Yes & 0.0065 \\
    SCAN & No\textsuperscript{†} & --- & Yes & 0.0072 \\
    \bottomrule
    \end{tabular}
    \footnotesize{\textsuperscript{†} MCMC runs from different starting configurations lead to different equilibrium distributions}
\end{threeparttable}
\end{table}

Table~\ref{tab:synthetic} summarises the results of these experiments.
Our MCMC for SCAN on the smaller $D/T=100$ dataset did not converge, whereas we get good convergence on the larger $D/T=500$ dataset. DiSC estimates a smaller number of parameters than SCAN, with lower variance but potential bias. This tradeoff seems to be advantageous for sense-labelling: as evidenced by the Brier scores on the larger dataset, DiSC sense-labelling on SCAN-friendly synthetic data is as good or better than SCAN itself.

The Brier score is a proper scoring rule, so we expect SCAN to do better on average over synthetic datasets --- on average but not necessarily in probability. Posterior plots similar to Figure~\ref{fig:bank_commercial_error_bars} (not included in this paper) confirm our intuition that, despite the large proportion of context words with a significant sense-time interaction effect, very few of these words appear with high probability across more than one sense. This may explain the Brier score ordering. It is possible to construct data for which SCAN scores more highly than DiSC by artificially fixing the prior parameters to include a very large proportion of words with \textit{both} high frequency in more than one sense \textit{and} strong interactions (cf. Section~\ref{sec:sense_change}). However, this seems unrepresentative of typical real data. In experiments where we explicitly introduce such strong interactions across multiple senses (discussed in Appendix~\ref{sec:appendix:synthetic}), we find that the performance of SCAN is a little better than that of DiSC but the gain is slight. In our experience, modelling sense-time interactions is detrimental for automatic sense-labelling.

\section{Experiment 4: Analysis of sense change for "kosmos"}
\label{sec:experiment_kosmos}

We now come to the principal application. We will make this analysis twice: here, and in Appendix~\ref{sec:appendix:kosmos}, depending on whether we exclude (as here) or include non-collocates. All translations of Greek words in this section have been obtained from Wiktionary.

The ancient Greek data for target word "kosmos" (\texttt{\textgreek{κόσμος}}) contains considerably fewer snippets than the "bank" data (cf. Section~\ref{sec:data}) whilst using almost the same sized vocabulary, making it a relatively sparse and noisy dataset. The "kosmos" data contains other features making this analysis harder than the "bank" analysis: the three senses of "kosmos" are not as well separated as those of "bank", and a number of context words appear with high probability under more than one sense in the expert annotation $o_d, d \in \{1,\dots,D\}$. Examples include \texttt{\textgreek{θεός}} (divine/god $\{0.26, 0.10, 0.64\}$), \texttt{\textgreek{ἀνήρ}} (man $\{0.45, 0.47, 0.08\}$) and \texttt{\textgreek{πόλις}} (city $\{0.30, 0.64, 0.06\}$) among others where, given that word $v$ appears as context in a snippet, it appears under sense $k$ with empirical probability $ \frac{N_{v,k,\cdot}^{W,o}}{N_{\cdot,k,\cdot}^{W,o}} \Big/ \sum_{l=1}^K \frac{N_{v,l,\cdot}^{W,o}}{N_{\cdot,l,\cdot}^{W,o}} $ for $k = \{1,\dots,K\}$, where $N_{v,k,\cdot}^{W,z} = \sum_{t=1}^T N_{v,k,t}^{W,z}$ and $N_{\cdot,k,\cdot}^{W,z} = \sum_{v=1}^V N_{v,k,\cdot}^{W,z}$. This makes the task of automatic sense-identification in the "kosmos" data particularly challenging.

\begin{table}
\caption{\label{tab:kosmos_confusion_matrix} "Kosmos" confusion matrix statistics using DiSC}
\centering
\begin{tabular}{l r r r}
    \toprule
                    & \multicolumn{3}{c}{Positive condition}\\
                    & Decoration  & Order  & World \\
    \midrule
    Sensitivity (true $+$ve) & 0.629 & 0.547 & 0.863 \\
    Specificity (true $-$ve) & 0.873 & 0.872 & 0.815 \\
    \bottomrule
\end{tabular}
\end{table}

Using our DiSC model with two genres, the words most probable a posteriori under each sense are as follows: \selectlanguage{greek}
\begin{Verbatim}[samepage=true,commandchars=\\\{\}]
1 γυνή  χρύσεος  φέρω     κοσμέω καλός σῶμα  ὅπλον ἐσθής   πλῆθος \textcolor{red}{τάξις}
2 πόλις πολιτεία πρότερος ἀνήρ   φέρω  νόμος μέγας πάρειμι ἀρχή   καθίστημι
3 γῆ    οὐρανός  ὅλος     θεός   εἷς   φύσις καλός σύμπας  ψυχή   σῶμα
\end{Verbatim} 
\selectlanguage{UKenglish}We compare these against ground truth. We identify the senses $\{1,2,3\}$ with the labels \{decoration, order, world\} by mapping our marginal posterior distributions $\frac{1}{T} \sum_{t=1}^T \tilde{\psi}^{k,t}$ for $k \in \{1,2,3\}$ to their closest empirical distributions $( N_{1,k,\cdot}^{W,o}, \dots, N_{V,k,\cdot}^{W,o} ) / N_{\cdot,k,\cdot}^{W,o}$ under the expert-annotated sense labels $o_d, d \in \{1,\dots,D\}$. We note that certain distinctive words that help identify the sense have been correctly assigned by our model, for example: \texttt{\textgreek{γυνή}} (woman) and \texttt{\textgreek{χρύσεος}} (golden) for "decoration"; \texttt{\textgreek{πολιτεία}} (citizenship) and \texttt{\textgreek{νόμος}} (custom/law) for "order"; \texttt{\textgreek{γῆ}} (earth) and \texttt{\textgreek{οὐρανός}} (sky) for "world". Some words such as \texttt{\textgreek{τάξις}} (literally "order" but assigned with "decoration") have been misplaced, although the error is small: if $v=\texttt{\textgreek{τάξις}}$ then $v$ is assigned by the model to sense $k$ with probability $p(z_d=k | v \in W_d, \tilde{\psi}) = \sum_{t=1}^T \tilde{\psi}^{k,t}_v \big/ \sum_{l=1}^K \sum_{t=1}^T \tilde{\psi}^{l,t}_v = \{0.43, 0.42, 0.15\}$ for $k = \{1,2,3\}$.

Using the Brier score to assess model performance, the baseline score under uniform random assignment is $BS = 0.67$ for $K=3$ senses. The realised score of $BS = 0.41$ using DiSC therefore indicates good model performance. Ignoring genre and setting $G=1$ gives $BS=0.47$, i.e. worse than the $G=2$ case, so the genre-covariate plays a role, in agreement with the views of \citet{GASC_2019arXiv190305587P}. As in the "bank" analysis, the confusion matrix from a similar classification task for the "kosmos" data is summarised in Table~\ref{tab:kosmos_confusion_matrix}, indicating generally good model performance. The relatively lower sensitivity for the "decoration" and "order" senses is reflective of the ambiguity discussed above.

\begin{figure}[!t]
\includegraphics[width = 1\textwidth, keepaspectratio]{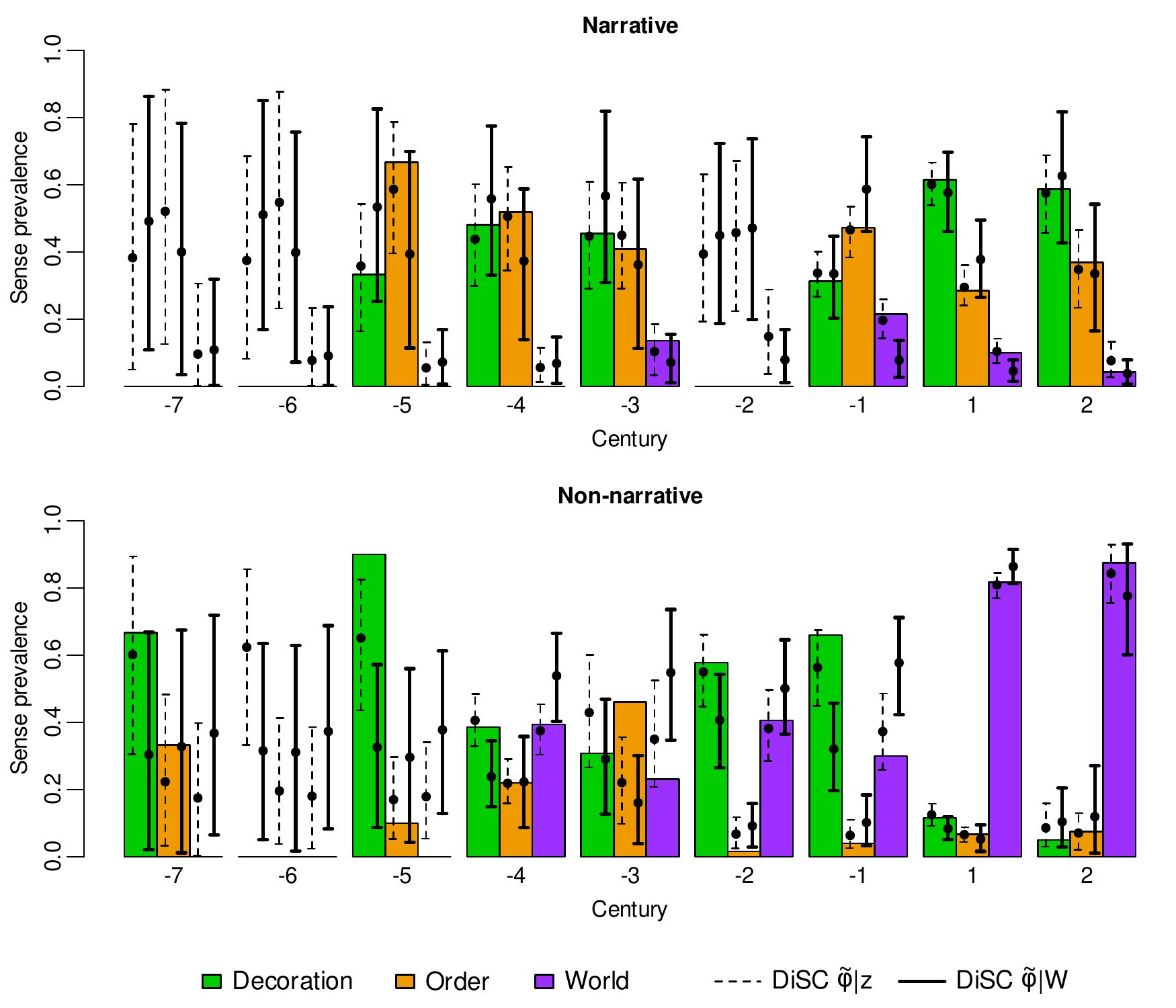}
\vspace*{-20pt}
\caption{"Kosmos" expert-annotated empirical sense prevalence (coloured bars with height $N_{k,g,t}^{o} / \sum_{l=1}^K N_{l,g,t}^{o}$ for each $k,g,t$) with 95\% HPD intervals (error bars) and posterior means (circles) from the DiSC output}
\label{fig:kosmos_error_bars}
\end{figure}

In Figure~\ref{fig:kosmos_error_bars} the 95\% HPD intervals for the marginal posteriors $\tilde{\phi}^{g,t}_k | W$ from the DiSC output have a high degree of overlap with the posteriors $\tilde{\phi}^{g,t}_k | (z=o)$ based on expert-annotated sense labels for all $k,g,t$, again indicating good model performance.
The DiSC HPD intervals contain the empirical estimates $N_{k,g,t}^{o} / \sum_{l=1}^K N_{l,g,t}^{o}$ for these data in most cases, with the exceptions being the time periods with little data (cf. Table~\ref{tab:kosmos_snippet_counts}). For these time periods, ground-truth sense prevalence estimates given by the posteriors  $\tilde{\phi}^{g,t}_k | (z=o)$ seem to us a more appropriate basis for comparison than the simple empirical estimates. Adjacent temporal data smooths these posteriors. Also, where there is limited data at both time $t$ and times $t \pm 1$, the increased uncertainty in the ground-truth is quantified in the wider HPD intervals for those times $t$. The very high degree of overlap between the marginal posteriors $\tilde{\phi}^{g,t}_k | W$ and those for $\tilde{\phi}^{g,t}_k | (z=o)$ indicates we are doing as well, for prevalence estimation with the unlabelled data, as we would do if we actually had the ground truth. 

Using GASC, on the other hand, we find that all MCMC fail to converge to the same stationary distribution starting from different random configurations. Exploration showed that the posterior distribution for GASC contains multiple metastable states with very long lifetimes and multiple modes (besides those associated with label switching). We do not see convergence even when using our best samplers with parallel tempering on multiple cores. This seems to be associated with over-parameterisation in the GASC model for this small and noisy dataset: the $\psi$ array has $VKT$ parameters in GASC compared to only $V(K+T)$ parameters in DiSC. This is more than double the number of parameters, and leads to multiple modes of near-equal fit as measured by the log-likelihood. 

The senses cannot be reliably identified from the model output as they are not sufficiently distinct. For example, our best run using GASC from six different random starting configurations gave the following most probable words: \selectlanguage{greek}
\begin{Verbatim}[samepage=true,commandchars=\\\{\}]
1 ἀνήρ \textcolor{red}{πολιτεία} ἀρχή  \textcolor{red}{γυνή}  κύριος  ἀξιόω \textcolor{red}{τάξις} πρότερος πάτριος  κόσμος
2 φέρω \textcolor{red}{γυνή}     πόλις καλός κοσμέω  θεός  μέγας \textcolor{red}{τάξις}    \textcolor{red}{πολιτεία} σῶμα  
3 γῆ   πόλις    ὅλος  θεός  οὐρανός εἷς   φύσις σῶμα     πέντε    σύμπας
\end{Verbatim} 
\selectlanguage{UKenglish}Since representative words for both "order" and "decoration" (such as \texttt{\textgreek{πολιτεία}} and \texttt{\textgreek{γυνή}}) appear with high probability under two senses, it is not clear how to assign the sense labels between them in order to make any sort of comparison. The same behaviour is seen further down the sense columns for other less frequent context words. Trying all possible permutations of the sense labels, the best Brier score we get is $BS = 0.73$ --- worse than randomly assigning the probability $\hat{p}(z_d=k) = \frac{1}{K}$ for all $d,k$.

\section{Conclusion}
\label{sec:conclusion}

We have given a generative model of Diachronic Sense Change (DiSC), treating sense and time as additive effects, and a gradient-based MCMC method for inferring model parameters. 

Whilst we adopt the overall modelling framework of \citet{frermann2016bayesian} and \citet{GASC_2019arXiv190305587P}, we found that the specific MCMC samplers used there take around 40 times as long to achieve the same ESS. Our MCMC exploits the fact that it is possible to marginalise by summing over the discrete sense assignments exactly. Gradient-based MCMC on the marginal is, not surprisingly, far more efficient than simple Gibbs sampling on the joint model. 

We carried out automatic sense-annotation by identifying word-sense dependence, and discovered sense groupings. We measured time-evolution of sense distributions over context words and of sense prevalence distributions over senses. We showed that, for the well-behaved "bank" data where MCMC targeting DiSC, SCAN and GASC all converge, DiSC has slightly better predictive performance for held-out expert-annotated sense labels despite being the simpler model. We further showed that, for the smaller and noisier "kosmos" data where no well-calibrated fitting procedure is available for SCAN and GASC, DiSC works well in the sense that it gives prevalence estimation intervals close to those we would obtain if we had the true sense labels. 

As far as we are aware, our analysis of "kosmos" in the test case Section~\ref{sec:experiment_kosmos} and "real-use" Appendix~\ref{sec:appendix:kosmos} give the first analysis of sense change for these data that returns good agreement with prevalence from expert annotation. One criticism examined in Section~\ref{sec:sense_change} is that DiSC does not model the sense-time interaction effect. In our setting, the sense of a target word is defined by the distribution over its context words. DiSC does model temporal change in this distribution, but captures only the main effects associated with changes in the overall usage frequency of the context words across all senses of the target word. It would be interesting to look for further examples of target words where the independent evolution of senses modelled by SCAN/GASC is measurable and impacts sense-labelling. 

Our study of synthetic data in Section~\ref{sec:experiment_synthetic} showed that even when data are simulated under SCAN, with abundant interaction, DiSC gives reliable sense-label predictions and parameter estimates. Furthermore, in many potentially interesting cases where the data are sparse, there is no computational procedure we know of that will actually fit the interaction as parameterised in SCAN/GASC reliably. In future work we would like to find a parameterisation of the sense-time interaction effect 
that is tractable. 
DiSC would then be a natural null model to this alternative.

All these models can identify the emergence of new senses, but we have not explored this. It may be possible to include an atom of probability at $\tilde\phi^{g,t}_k=0$ in order to carry out simultaneous and formal model-selection for the first time $t$ at which $\tilde{\phi}^{g,t}_k>0$ and there is evidence for a new sense.

DiSC shares some disadvantages with the generative models on which it is based. We target the senses of one word at a time, in contrast to some approaches in the machine learning literature cited in Section~\ref{sec:related_work}. However, DiSC does provide well-calibrated uncertainty estimates, as evidenced by our experiments, rather than just point estimates. Another limitation is that the number of senses $K$ needs to be checked in multiple runs. It may be possible to learn $K$ using more formal testing procedures. However, for estimation of sense distributions (over context words), sense prevalence and their evolution from unlabelled text, DiSC works well in a semi-supervised mode in which model selection is based on the requirement that the output be meaningful to the user.

\section*{Implementation}

R scripts and data files used to produce the results, figures and tables reported in the paper are available from \url{https://github.com/schyanzafar/DiSC}.

\section*{Acknowledgement}

This research is funded by the Engineering and Physical Sciences Research Council (EPSRC) under grant EP/S515541/1.

\begin{appendices}

\section{GASC generative model}
\label{sec:appendix:gasc_model}

The GASC generative model is given in Algorithm \ref{alg:GASC_generative_model}. Note:
\begin{enumerate}
    \item Following \citet{frermann2016bayesian}, the authors of \citet{GASC_2019arXiv190305587P} give generative prior processes for $\phi$ and $\psi$ in terms of the full conditional distributions. This is natural as there is an improper initialisation. In lines \ref{alg_line:gasc_priors_start}--\ref{alg_line:gasc_priors_end} we present the generative process as an equivalent Markov chain \citep{rue2005gaussian}. 
    \item In order to make an exact numerical match with GASC, $\kappa_\phi$ in Algorithm~\ref{alg:GASC_generative_model} has a factor of two multiplier so the $\kappa_\phi$ parameters in GASC and DiSC differ by a factor of two even when $\alpha_\phi=1$ in the AR(1) process.
    \item Lines \ref{alg_line:gasc_SWs1} and \ref{alg_line:gasc_SWs2} treat stopwords and uninformative words explicitly. Stopwords may be treated as punctuation and dropped without counting as occupying a context position. Also, \citet{GASC_2019arXiv190305587P} do not discuss how uninformative words or hapaxes are identified in snippets, so this is a change on their setup. We chose to treat these words in GASC as we do for DiSC as it seemed only an improvement.
\end{enumerate}

\begin{algorithm}[!t]
\caption{GASC: generative model}
\label{alg:GASC_generative_model}
\begin{algorithmic}[1]

\centering
\Statex ------------------------ PRIOR MODEL ------------------------ 

\justifying
\State fix hyperparameters $\kappa_{\psi}, a, b$

\State draw $\kappa_{\phi} \sim \InvGamma(a,b)$

\State initialise at time $t=1$ \label{alg_line:gasc_priors_start}
    \For {genre $g \in 1:G$}
        \State draw sense prevalence parameter from improper uniform $\phi^{g,1} \sim \pi(\phi^{g,1}) \propto 1$
    \EndFor
   \For {sense $k \in 1:K$}
        \State draw word parameter from improper uniform density $\psi^{k,1} \sim \pi(\psi^{k,1}) \propto 1$
    \EndFor 

\For {time $t \in 2:T$}
    \For {genre $g \in 1:G$}
        \State draw sense prevalence parameter $\phi^{g,t} | \phi^{g,t-1}, \kappa_{\phi} \sim \mathcal{N} \left( \phi^{g,t-1}, \diag ( 2\kappa_{\phi} ) \right)$
    \EndFor
    \For {sense $k \in 1:K$}
        \State draw word parameter $\psi^{k,t} | \psi^{k,t-1}, \kappa_{\psi} \sim \mathcal{N} \left( \psi^{k,t-1}, \diag ( 2 \kappa_{\psi} ) \right)$
    \EndFor 
\EndFor \label{alg_line:gasc_priors_end}

\State using softmax \eqref{eq:softmax}, transform real arrays $\phi$ and $\psi$ into probability arrays $\tilde{\phi}$ and $\tilde{\psi}$

\centering
\Statex ------------------------ OBSERVATION MODEL ------------------------ 

\justifying
\State fix probabilities of drawing stopwords $q^\text{SW}$ and uninformative words $q^\text{U}$ \label{alg_line:gasc_SWs1}

\For {snippet $d \in 1:D$}
    \State draw number of context words $L_d | L, q^\text{SW}, q^\text{U} \sim \Bin(L, 1 - q^\text{SW} - q^\text{U})$ \label{alg_line:gasc_SWs2}
    \State draw a random subset $\{ i_1 , \dots, i_{L_d} \}$ of size $L_d$ from $\{1, \dots, L\}$
    \State draw sense assignment $z_{d} | \tilde{\phi}^{\gamma_d,\tau_d} \sim \Mult \left( \tilde{\phi}^{\gamma_d,\tau_d}_{1}, \dots, \tilde{\phi}^{\gamma_d,\tau_d}_{K} \right)$
    \For {context position $i \in \{ i_1 , \dots, i_{L_d} \}$}
        \State draw context word $w_{d,i} | z_d, \tilde{\psi}^{z_d,\tau_d} \sim \Mult \left( \tilde{\psi}^{z_d,\tau_d}_{1}, \dots, \tilde{\psi}^{z_d,\tau_d}_{V} \right)$
    \EndFor
\EndFor

\end{algorithmic}
\end{algorithm}

\section{Gibbs samplers}
\label{sec:appendix:existing_methods}

We briefly describe and critique current MCMC methods for sampling the posterior $\phi^{g,t} | z$ in the DiSC and GASC models. Sampling $\psi^{k,t} | z, W$ in the GASC model is very similar. The conditional prior distribution is
\begin{align} \label{eq:phi_prior_conditionals}
    \phi^{g,t} | \phi^{g,-t} \sim 
    \begin{cases}
        \mathcal{N} \left( \alpha_\phi \phi^{g,t+1}, \diag ( \kappa_{\phi} ) \right) &\text{ if } t=1 \\
        \mathcal{N} \left( \frac{\alpha_\phi} {1+(\alpha_\phi)^2} (\phi^{g,t-1} + \phi^{g,t+1}), \diag \left( \frac{\kappa_\phi}{1+(\alpha_\phi)^2} \right) \right) &\text{ if } t \in 2:(T-1) \\
        \mathcal{N} \left( \alpha_\phi \phi^{g,t-1}, \diag ( \kappa_{\phi} ) \right) &\text{ if } t=T 
    \end{cases}
\end{align}
under DiSC. For GASC, we replace $\kappa_{\phi}$ with $2\kappa_{\phi}$ and substitute $\alpha_\phi = 1$.

\subsection{Auxiliary uniform variable method}
\label{sec:appendix:auxiliary_uniform}

The auxiliary uniform variable method of \citet{mimno2008gibbs}, which is based on the method of \citet{GROENEWALD2005857}, is used to sample $\phi_k^{g,t} | \phi_{-k}^{g,t}, z$ iteratively over $k \in 1:K$ by moving within a weighted and bounded region, where the weights are determined by the prior distributions and the bounds are determined by $N_{k,g,t}^{z}$.

\begin{figure}
\centering
\includegraphics[width = 1\textwidth, keepaspectratio]{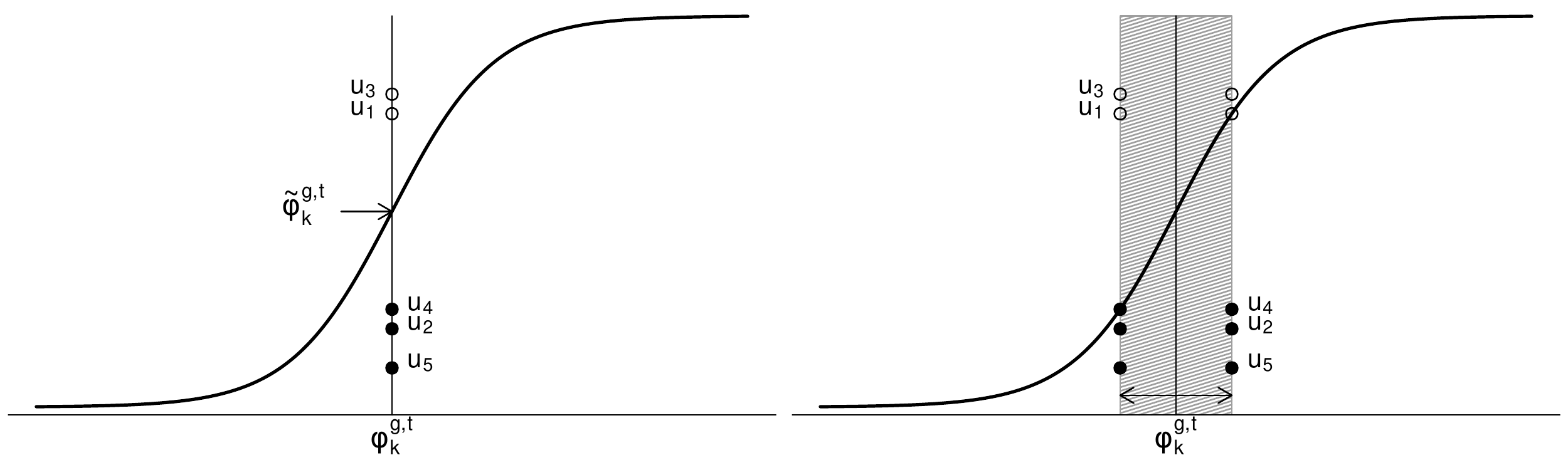}
\vspace*{-20pt}
\caption{Relationship between $\phi_k^{g,t}$, $z$, and $u$ in the case of 5 snippets. LEFT: Given $\phi_k^{g,t}$, if $z_d=k$ then $u_d$ (black circle) falls below $\tilde{\phi}_k^{g,t}$; and if $z_d \neq k$ then $u_d$ (white circle) falls above $\tilde{\phi}_k^{g,t}$. RIGHT: With $u$ given, $\phi_k^{g,t}$ can be anywhere within the interval defined by the highest black circle and the lowest white circle. Figure adapted from \citet{mimno2008gibbs}.}
\label{fig:Logistic_normal_sampling}
\end{figure}

Figure \ref{fig:Logistic_normal_sampling} shows an example. Given $N_{k,g,t}^{z}$ snippets assigned sense $k$ and $N_{\cdot,g,t}^{z}$ snippets in total for time $t$ and genre $g$, we draw $N_{k,g,t}^{z}$ uniform variables below the logistic CDF and $N_{\cdot,g,t}^{z} - N_{k,g,t}^{z}$ uniform variables above the logistic CDF. That is, for all $d \in \{d': \tau_{d'}=t \text{ and } \gamma_{d'}=g \}$ we draw 
\begin{align}
    u_d \sim 
    \begin{cases}
        \mathcal{U} (0, \tilde{\phi}_k^{g,t}) &\text{ if } z_d = k \\
        \mathcal{U} (\tilde{\phi}_k^{g,t}, 1) &\text{ if } z_d \neq k \text{,}
    \end{cases}
\end{align}
and sample a new $\phi_k^{g,t}$ from the conditional prior \eqref{eq:phi_prior_conditionals} truncated to the bounded region
\begin{equation}\label{eq:logistic_normal_bounds}
\max_{d:z_d=k} \log \frac{C u_d}{1-u_d} < \phi_k^{g,t} < \min_{d:z_d \neq k} \log \frac{C u_d}{1-u_d}    
\end{equation}
where
\begin{equation}\label{eq:C_bound_factor}
C = \sum_{k' \neq k} \exp(\phi_{k'}^{g,t}).
\end{equation}
The efficiency of this procedure can be improved by sampling two Beta variables instead of $N_{\cdot,g,t}^{z}$ uniform variables.

The problem with this method is that the bounded region \eqref{eq:logistic_normal_bounds} can be very narrow whenever the dimensions $N_{k,g,t}^{z}$ and $N_{\cdot,g,t}^{z}$ are large, leading to a very small move in each iteration. This is because sampling a large number of uniform variables below or above the CDF is likely to result in at least one variable being close to the curve. The resulting convergence is therefore very slow.

\subsection{Auxiliary Polya-Gamma variable method}
\label{sec:appendix:auxiliary_pg}

The auxiliary Polya-Gamma variable method of \citet{PolyaGamma_2012arXiv1205.0310P} is used to sample $\phi_k^{g,t} | \phi_{-k}^{g,t}, z$ iteratively over $k \in 1:K$ by first drawing
\begin{equation}
    \omega \sim \mathcal{PG}(N_{\cdot,g,t}^{z}, \eta) \text{,}
\end{equation}
from the Polya-Gamma distribution $\mathcal{PG}$, where $\eta = \phi_k^{g,t} - \log C$ with $C$ as in \eqref{eq:C_bound_factor}, and then sampling 
\begin{equation}
    \phi_k^{g,t} \sim \mathcal{N} (m,\Sigma) \text{,}
\end{equation}
where $\Sigma = (\sigma^{-2} + \omega)^{-1}$, $m = \Sigma ( \mu \sigma^{-2} + N_{k,g,t}^{z} - \frac{1}{2} N_{\cdot,g,t}^{z} + \omega \log C )$, and $\mu$ and $\sigma^2$ are the mean and variance for the relevant conditional prior distribution defined in \eqref{eq:phi_prior_conditionals} and depending on $t$.

A draw from a Polya-Gamma distribution is computationally expensive whenever the shape parameter $N_{\cdot,g,t}^{z}$ is large. This is very often the case with any real dataset, especially for $\psi$ updates. \citet{chen2013scalable} give an alternative approximate method which cuts the computational cost of each draw from $\mathcal{O}(N_{\cdot,g,t}^{z})$ down to $\mathcal{O}(1)$. They note that if $x_i \sim \mathcal{PG}(1, \eta)$ then $\omega = \sum_{i=1}^N x_i \sim \mathcal{PG}(N, \eta)$ by the additive property of the Polya-Gamma distribution. Therefore, by the central limit theorem, $\omega$ is approximately Gaussian for large $N$, and may be obtained by transforming another approximately Gaussian random variable $\lambda \sim \mathcal{PG}(M, \eta)$ viz.
\begin{equation}
    \sqrt{\Var(\omega)/Var(\lambda)} (\lambda - \mathbb{E}[\lambda]) + \mathbb{E}[\omega]
\end{equation}
where $\mathbb{E}[\omega] = \frac{N}{2\eta} \tanh{\frac{\eta}{2}}$ and $\Var(\omega)/Var(\lambda) = N/M$. This approximate strategy works well even when $M=1$. The method is relatively simple. However, parameter inference is not asymptotically exact and, as we saw in Section~\ref{sec:experiment_samplers}, the method is in any case dominated by a hybrid MALA-HMC sampler which is asymptotically exact.

\section{Gradient-based MCMC methods}
\label{sec:appendix:new_method}

We sample the conditional posteriors \eqref{eq:phi_posterior}--\eqref{eq:chi_posterior} for each variable by proposing an update from a suitable distribution and accepting or rejecting it using the Hastings ratio, iterating over the columns. For example for $\phi$, we propose a candidate vector
\begin{equation}
    \phi^{*g,t} | \phi^{g,\cdot}, \psi^{\cdot,t}, W_{\mathcal{D}(g,t)} \sim q(\cdot | \phi^{g,\cdot}, \psi^{\cdot,t}, W_{\mathcal{D}(g,t)})
\end{equation}
and accept it with probability
\begin{equation} \label{eq:phi_MH_ratio}
    1 \wedge \frac{\pi(\phi^{*g,t} | \phi^{g,-t}) p(W_{\mathcal{D}(g,t)} | \phi^{*g,t}, \psi^{\cdot,t})} {\pi(\phi^{g,t} | \phi^{g,-t}) p(W_{\mathcal{D}(g,t)} | \phi^{g,t}, \psi^{\cdot,t})} \frac{q(\phi^{g,t}|\phi^{*\cdot,g}, \psi^{\cdot,t}, W_{\mathcal{D}(g,t)})} {q(\phi^{*g,t}|\phi^{g,\cdot}, \psi^{\cdot,t}, W_{\mathcal{D}(g,t)})} 
\end{equation}
where $p(W_{\mathcal{D}(g,t)} | \phi^{g,t}, \psi^{\cdot,t})$ is the likelihood
\begin{equation} \label{eq:likelihood_tg}
    p(W_{\mathcal{D}(g,t)} | \phi^{g,t}, \psi^{\cdot,t}) = \prod_{d \in \mathcal{D}(g,t)} \sum_{k=1}^K \tilde{\phi}_k^{\gamma_d,\tau_d} \prod_{i=i_1}^{i_{L_d}} \tilde{\psi}_{w_{d,i}}^{k,\tau_d}
\end{equation}
and $\mathcal{D}(g,t) = \{ d: \gamma_d \in g \text{ and } \tau_d \in t \}$ is the set of snippet indices for time(s) $t$ and genre(s) $g$. Updates for $\theta, \chi$ for DiSC or $\psi$ for GASC are analogous. The conditional prior density $\pi(\phi^{g,t} | \phi^{g,-t})$ is defined by \eqref{eq:phi_prior_conditionals}, whereas for $\theta$ we have the conditional
\begin{align} \label{eq:theta_prior_conditionals}
    \theta^{t} | \theta^{-t} \sim 
    \begin{cases}
        \mathcal{N} \left( \alpha_\theta \theta^{t+1}, \diag ( \kappa_{\theta} ) \right) &\text{ if } t=1 \\
        \mathcal{N} \left( \frac{\alpha_\theta} {1+(\alpha_\theta)^2} (\theta^{t-1} + \theta^{t+1}), \diag \left( \frac{\kappa_\theta}{1+(\alpha_\theta)^2} \right) \right) &\text{ if } t \in 2:(T-1) \\
        \mathcal{N} \left( \alpha_\theta \theta^{t-1}, \diag ( \kappa_{\theta} ) \right) &\text{ if } t=T \text{,}
    \end{cases}
\end{align}
and for $\chi$ we have, unconditionally,
\begin{equation} \label{eq:chi_prior}
    \chi^k \sim \mathcal{N} \left( 0, \diag(\kappa_{\chi}) \right) \text{.}
\end{equation}
In practice, it may be more efficient to update all columns of $\chi$ together, which is the approach we take in our R implementation for the applications in this paper.

The proposal density $q(\cdot | \phi^{g,\cdot}, \psi^{\cdot,t}, W_{\mathcal{D}(g,t)})$ is constructed using gradient-based methods such as MALA \citep{10.2307/3318418} or HMC \citep{1987PhLB..195..216D}, which make proposals in the direction where the posterior density is increasing. MALA, for instance, uses the proposal distribution 
\begin{equation} \label{eq:proposal_distribution}
    \phi^{*g,t} | \phi^{g,\cdot}, \psi^{\cdot,t}, W_{\mathcal{D}(g,t)} \sim \mathcal{N} \left( \phi^{g,t} + \frac{\sigma_\phi^2}{2} \nabla_{\phi^{g,t}} \log \pi(\phi^{g,t} | \phi^{g,-t}, \psi^{\cdot,t}, W_{\mathcal{D}(g,t)} ) , \sigma_\phi^2 \Sigma_\phi \right)
\end{equation}
for $\phi$ updates, and analogous distributions for the other variables, where $\sigma_x^2$ and $\Sigma_x$ are the proposal scale and covariance parameters for variable $x$ respectively. We keep the covariance $\Sigma_x$ fixed at the appropriate identity matrix, whereas we tune the scale $\sigma_x^2$ using the log-adaptive proposals of \citet{shaby2010exploring}. When using HMC with the leapfrog method (see e.g. \citet{2012arXiv1206.1901N}), we keep the number of leapfrog steps fixed and use the same technique to tune the leapfrog step size $\sigma_x^2$. We experimented with tuning the covariance $\Sigma_x$, including varying diagonal elements, without gain.

The tuning is done by calculating the running empirical acceptance rate $\Bar{\alpha} = \frac{\text{\# jumps}}{N}$ based on the last $N$ MCMC iterations, and adjusting the scale parameter up or down via $\log \sigma_x^2 \leftarrow \log \sigma_x^2 + C_n (\Bar{\alpha} - \alpha^\text{opt})$, where $C_n$ is a parameter that decreases with iteration number $n$ (so that $\sigma_x^2$ tends to a constant as $n \rightarrow \infty$) and $\alpha^\text{opt}$ is the target optimal acceptance rate. Under certain conditions, optimal asymptotic acceptance rates have been proposed as 0.574 for MALA \citep{doi:10.1111/1467-9868.00123} and 0.651 for HMC \citep{beskos2013}, which are the values we use for $\alpha^\text{opt}$ in our implementations.

The gradient of the log posterior density in \eqref{eq:proposal_distribution} can be broken up into the sum of gradients of the log prior density and the log likelihood. The gradients of the log prior densities $\nabla_{\phi^{g,t}} \log \pi(\phi^{g,t} | \phi^{g,-t})$, $\nabla_{\theta^t} \log \pi(\theta^{t} | \theta^{-t})$ and $\nabla_{\chi^k} \log \pi(\chi^k)$ are of the form $-V_x^{-1}(x-\mu_x)$ where the mean vector $\mu_x$ and covariance matrix $V_x$ for variable $x$ are given by \eqref{eq:phi_prior_conditionals}, \eqref{eq:theta_prior_conditionals} and \eqref{eq:chi_prior} respectively. The gradients of the log likelihoods are derived in the next subsection.

\begin{figure}[!t]
\centering
\includegraphics[width = 0.9\textwidth, keepaspectratio]{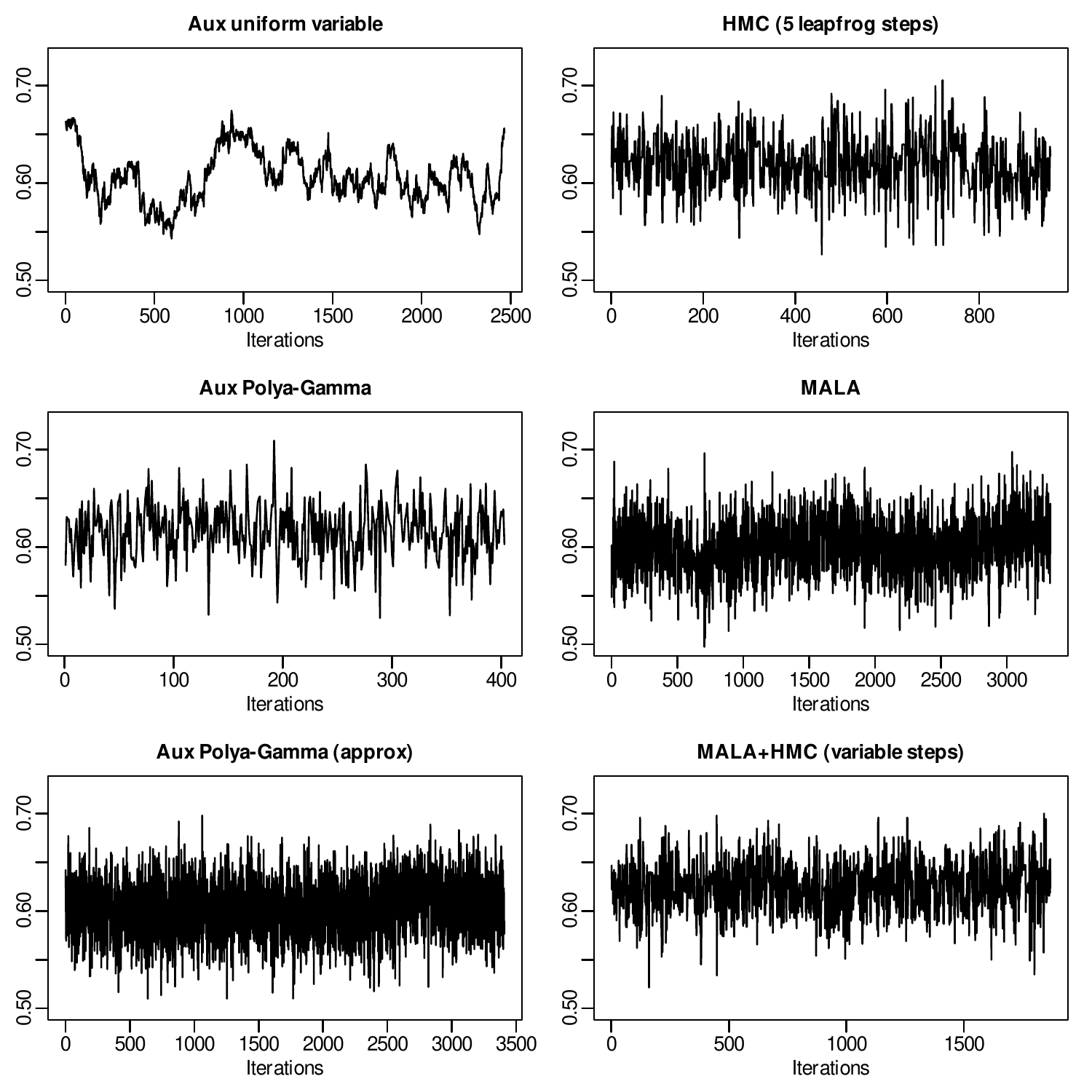}
\caption{$\tilde{\phi}_k^{g,t}$ trace plots with equal run times (after burn-in) for the institution sense of "bank" at time 1850-70 from the unlabelled SCAN posterior}
\label{fig:trace_plots}
\end{figure}

As we find in Section~\ref{sec:experiment_samplers} above, these gradient-based methods give significant efficiency gains over methods like the auxiliary uniform or the (asymptotically exact) auxiliary Polya-Gamma samplers described in Appendix~\ref{sec:appendix:existing_methods}. This is illustrated in Figure~\ref{fig:trace_plots} in which the $x$-axis shows the same elapsed time in CPU seconds across all plots and the $y$-axis shows the MCMC state for the prevalence parameter in one genre and time period. The gradient-based samplers shown on the right give much more rapid mixing than the other two methods. 

\subsection{Derivation of $\nabla_{\phi^{g,t}} \log p(W_{\mathcal{D}(g,t)} | \phi^{g,t}, \psi^{\cdot,t})$}

From equation \eqref{eq:likelihood_tg} we get 
\begin{equation*}
    \log p(W_{\mathcal{D}(g,t)} | \phi^{g,t}, \psi^{\cdot,t}) = \sum_{d \in \mathcal{D}(g,t)} \log \sum_{k=1}^K \tilde{\phi}_k^{\gamma_d,\tau_d} \prod_{i=i_1}^{i_{L_d}} \tilde{\psi}_{w_{d,i}}^{k,\tau_d} \text{,}
\end{equation*}
and taking derivatives with respect to $\phi^{g,t}_j$ gives
\begin{equation} \label{eq:diff_log_p_phi}
    \frac{\partial}{\partial \phi^{g,t}_j} \log p(W_{\mathcal{D}(g,t)} | \phi^{g,t}, \psi^{\cdot,t}) = \sum_{d \in \mathcal{D}(g,t)} \frac {\frac{\partial}{\partial \phi^{g,t}_j} \sum_{k=1}^K \tilde{\phi}_k^{g,t} \prod_{i=i_1}^{i_{L_d}} \tilde{\psi}_{w_{d,i}}^{k,t} } {\sum_{k=1}^K \tilde{\phi}_k^{g,t} \prod_{i=i_1}^{i_{L_d}} \tilde{\psi}_{w_{d,i}}^{k,t}} \text{.}
\end{equation}
Now, $\tilde{\phi}^{g,t}_k = \frac {\exp (\phi_k^{g,t})} {\sum_{k'=1}^K \exp (\phi_{k'}^{g,t})}$ so that 
\begin{equation} \label{eq:diff_phi_tilde}
    \frac{\partial \tilde{\phi}^{g,t}_k}{\partial \phi^{g,t}_j}
        = \frac {\exp(\phi_j^{g,t}) \mathbb{I}(j=k) \sum_{k'=1}^K \exp (\phi_{k'}^{g,t}) - \exp(\phi_j^{g,t}) \exp(\phi_k^{g,t}) } {\left( \sum_{k'=1}^K \exp (\phi_{k'}^{g,t}) \right)^2}
        = \tilde{\phi}^{g,t}_j \left( \mathbb{I}(j=k) - \tilde{\phi}^{g,t}_k \right) \text{.}
\end{equation}
Hence we have
\begin{equation*}
    \frac{\partial}{\partial \phi^{g,t}_j} \sum_{k=1}^K \tilde{\phi}_k^{g,t} \prod_{i=i_1}^{i_{L_d}} \tilde{\psi}_{w_{d,i}}^{k,t} = \tilde{\phi}_j^{g,t} \left( \prod_{i=i_1}^{i_{L_d}} \tilde{\psi}_{w_{d,i}}^{j,t} - \sum_{k=1}^K \tilde{\phi}_k^{g,t} \prod_{i=i_1}^{i_{L_d}} \tilde{\psi}_{w_{d,i}}^{k,t} \right) \text{,}
\end{equation*}
and substituting this into \eqref{eq:diff_log_p_phi} gives
\begin{equation}
     \frac{\partial}{\partial \phi^{g,t}_j} \log p(W_{\mathcal{D}(g,t)} | \phi^{g,t}, \psi^{\cdot,t}) = \sum_{d \in \mathcal{D}(g,t)} \frac {\tilde{\phi}_j^{g,t} \prod_{i=i_1}^{i_{L_d}} \tilde{\psi}_{w_{d,i}}^{j,t} } {\sum_{k=1}^K \tilde{\phi}_k^{g,t} \prod_{i=i_1}^{i_{L_d}} \tilde{\psi}_{w_{d,i}}^{k,t}} - \tilde{\phi}_j^{g,t} N_{\cdot,g,t}^{z}
\end{equation}
which are the elements of vector $\nabla_{\phi^{g,t}} \log p(W_{\mathcal{D}(g,t)} | \phi^{g,t}, \psi^{\cdot,t})$ for $j \in \{1,\dots,K\}$.

\subsection{Derivation of $\nabla_{\psi^{k,t}} \log p(W_{\mathcal{D}(1:G,t)} | \phi^{\cdot,t}, \psi^{\cdot,t})$}

The gradient vector $\nabla_{\psi^{k,t}} \log p(W_{\mathcal{D}(1:G,t)} | \phi^{\cdot,t}, \psi^{\cdot,t})$ is required for inferring $\psi$ in the GASC model. Differentiating the log likelihood with respect to $\psi^{k,t}_j$ gives 
\begin{equation} \label{eq:diff_log_p_psi}
    \frac{\partial}{\partial \psi^{k,t}_j} \log p(W_{\mathcal{D}(1:G,t)} | \phi^{\cdot,t}, \psi^{\cdot,t}) = \sum_{d \in \mathcal{D}(1:G,t)} \frac {\frac{\partial}{\partial \psi^{k,t}_j} \sum_{l=1}^K \tilde{\phi}_l^{\gamma_d,t} \prod_{i=i_1}^{i_{L_d}} \tilde{\psi}_{w_{d,i}}^{l,t} } {\sum_{l=1}^K \tilde{\phi}_l^{\gamma_d,t} \prod_{i=i_1}^{i_{L_d}} \tilde{\psi}_{w_{d,i}}^{l,t}} \text{.}
\end{equation}
As in \eqref{eq:diff_phi_tilde}, we have $\frac{\partial \tilde{\psi}^{l,t}_v}{\partial \psi^{k,t}_j} = \tilde{\psi}^{l,t}_v \left( \mathbb{I}(j=v) - \tilde{\psi}^{l,t}_j \right)$ if $k=l$, and $\frac{\partial \tilde{\psi}^{l,t}_v}{\partial \psi^{k,t}_j} = 0$ otherwise. Hence we have
\begingroup
\allowdisplaybreaks
\begin{align*}
    \frac{\partial}{\partial \psi^{k,t}_j} \sum_{l=1}^K \tilde{\phi}_l^{\gamma_d,t} \prod_{i=i_1}^{i_{L_d}} \tilde{\psi}_{w_{d,i}}^{l,t} 
    &= \frac{\partial}{\partial \psi^{k,t}_j} \tilde{\phi}_k^{\gamma_d,t} \prod_{i=i_1}^{i_{L_d}} \tilde{\psi}_{w_{d,i}}^{k,t} \\
    &= \tilde{\phi}_k^{\gamma_d,t} \sum_{i=i_1}^{i_{L_d}} \tilde{\psi}_{w_{d,i}}^{k,t} \left( \mathbb{I}(j=w_{d,i}) - \tilde{\psi}_{j}^{k,t} \right) \prod_{i' \neq i} \tilde{\psi}_{w_{d,i'}}^{k,t} \\
    &= \tilde{\phi}_k^{\gamma_d,t} \sum_{i=i_1}^{i_{L_d}} \left( \mathbb{I}(j=w_{d,i}) - \tilde{\psi}_{j}^{k,t} \right) \prod_{i'=i_1}^{i_{L_d}} \tilde{\psi}_{w_{d,i'}}^{k,t} \\
    &= \left( \tilde{\phi}_k^{\gamma_d,t} \prod_{i=i_1}^{i_{L_d}} \tilde{\psi}_{w_{d,i}}^{k,t} \right) \left( \sum_{i=i_1}^{i_{L_d}} \mathbb{I}(j=w_{d,i}) - L_d \tilde{\psi}_{j}^{k,t} \right) 
\end{align*}
\endgroup
and substituting this into \eqref{eq:diff_log_p_psi} gives
\begin{equation} \label{eq:diff_log_p_psi_final}
    \frac{\partial}{\partial \psi^{k,t}_j} \log p(W_{\mathcal{D}(1:G,t)} | \phi^{\cdot,t}, \psi^{\cdot,t}) = \sum_{d \in \mathcal{D}(1:G,t)} \frac {\tilde{\phi}_k^{\gamma_d,t} \prod_{i=i_1}^{i_{L_d}} \tilde{\psi}_{w_{d,i}}^{k,t}} {\sum_{l=1}^K \tilde{\phi}_l^{\gamma_d,t} \prod_{i=i_1}^{i_{L_d}} \tilde{\psi}_{w_{d,i}}^{l,t}} \left( \sum_{i=i_1}^{i_{L_d}} \mathbb{I}(j=w_{d,i}) - L_d \tilde{\psi}_{j}^{k,t} \right) 
\end{equation}
which are the elements of vector $\nabla_{\psi^{k,t}} \log p(W_{\mathcal{D}(1:G,t)} | \phi^{\cdot,t}, \psi^{\cdot,t})$ for $j \in \{1,\dots,V\}$.

\subsection{Derivation of $\nabla_{\theta^t} \log p(W_{\mathcal{D}(1:G,t)} | \phi^{\cdot,t}, \psi^{\cdot,t})$}

The relationship $\psi^{k,t}_j = \chi^k_j + \theta^t_j$ gives $\frac{\partial \psi^{k,t}_j}{\partial \theta^t_j} = 1$ for all $k \in \{1,\dots,K\}$, so applying the chain rule to \eqref{eq:diff_log_p_psi_final} we get
\begin{align}
    \frac{\partial}{\partial \theta^t_j} \log p(W_{\mathcal{D}(1:G,t)} | \phi^{\cdot,t}, \psi^{\cdot,t}) 
    &= \sum_{k=1}^K \frac{\partial}{\partial \psi^{k,t}_j} \log p(W_{\mathcal{D}(1:G,t)} | \phi^{\cdot,t}, \psi^{\cdot,t}) \nonumber\\
    &= \sum_{d \in \mathcal{D}(1:G,t)} \left( \sum_{i=i_1}^{i_{L_d}} \mathbb{I}(j=w_{d,i}) - L_d \sum_{k=1}^K \frac {\tilde{\phi}_k^{\gamma_d,t} \prod_{i=i_1}^{i_{L_d}} \tilde{\psi}_{w_{d,i}}^{k,t}} {\sum_{l=1}^K \tilde{\phi}_l^{\gamma_d,t} \prod_{i=i_1}^{i_{L_d}} \tilde{\psi}_{w_{d,i}}^{l,t}} \tilde{\psi}_{j}^{k,t} \right)
\end{align}
which are the elements of vector $\nabla_{\theta^t} \log p(W_{\mathcal{D}(1:G,t)} | \phi^{\cdot,t}, \psi^{\cdot,t})$ for $j \in \{1,\dots,V\}$.

\subsection{Derivation of $\nabla_{\chi^k} \log p(W | \phi, \psi)$}

The relationship $\psi^{k,t}_j = \chi^k_j + \theta^t_j$ gives $\frac{\partial \psi^{k,t}_j}{\partial \chi^k_j} = 1$ for all $t \in \{1,\dots,T\}$, so given the independence between time periods and applying the chain rule to \eqref{eq:diff_log_p_psi_final} we get
\begin{align}
     \frac{\partial}{\partial \chi^k_j} \log p(W | \phi, \psi) 
     &= \sum_{t=1}^T \frac{\partial}{\partial \psi^{k,t}_j} \log p(W_{\mathcal{D}(1:G,t)} | \phi^{\cdot,t}, \psi^{\cdot,t}) \nonumber\\
     &= \sum_{d=1}^D \frac {\tilde{\phi}_k^{\gamma_d,\tau_d} \prod_{i=i_1}^{i_{L_d}} \tilde{\psi}_{w_{d,i}}^{k,\tau_d}} {\sum_{l=1}^K \tilde{\phi}_l^{\gamma_d,\tau_d} \prod_{i=i_1}^{i_{L_d}} \tilde{\psi}_{w_{d,i}}^{l,\tau_d}} \left( \sum_{i=i_1}^{i_{L_d}} \mathbb{I}(j=w_{d,i}) - L_d \tilde{\psi}_{j}^{k,\tau_d} \right) 
\end{align}
which are the elements of vector $\nabla_{\chi^k} \log p(W | \phi, \psi)$ for $j \in \{1,\dots,V\}$.

\section{"Bank" additional results}
\label{sec:appendix:bank}

\begin{figure}[!t]
\includegraphics[width = 1\textwidth, keepaspectratio]{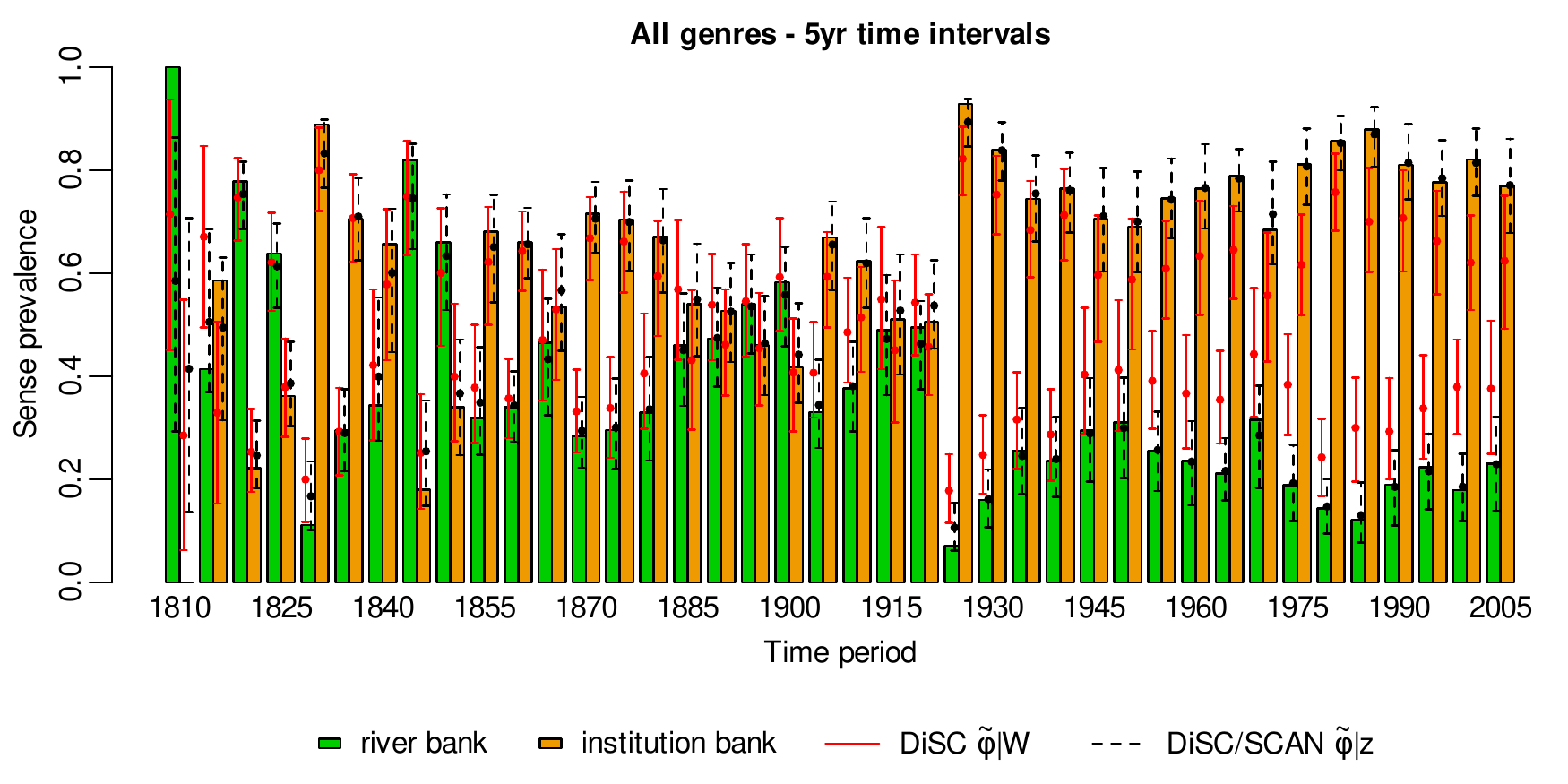}
\includegraphics[width = 1\textwidth, keepaspectratio]{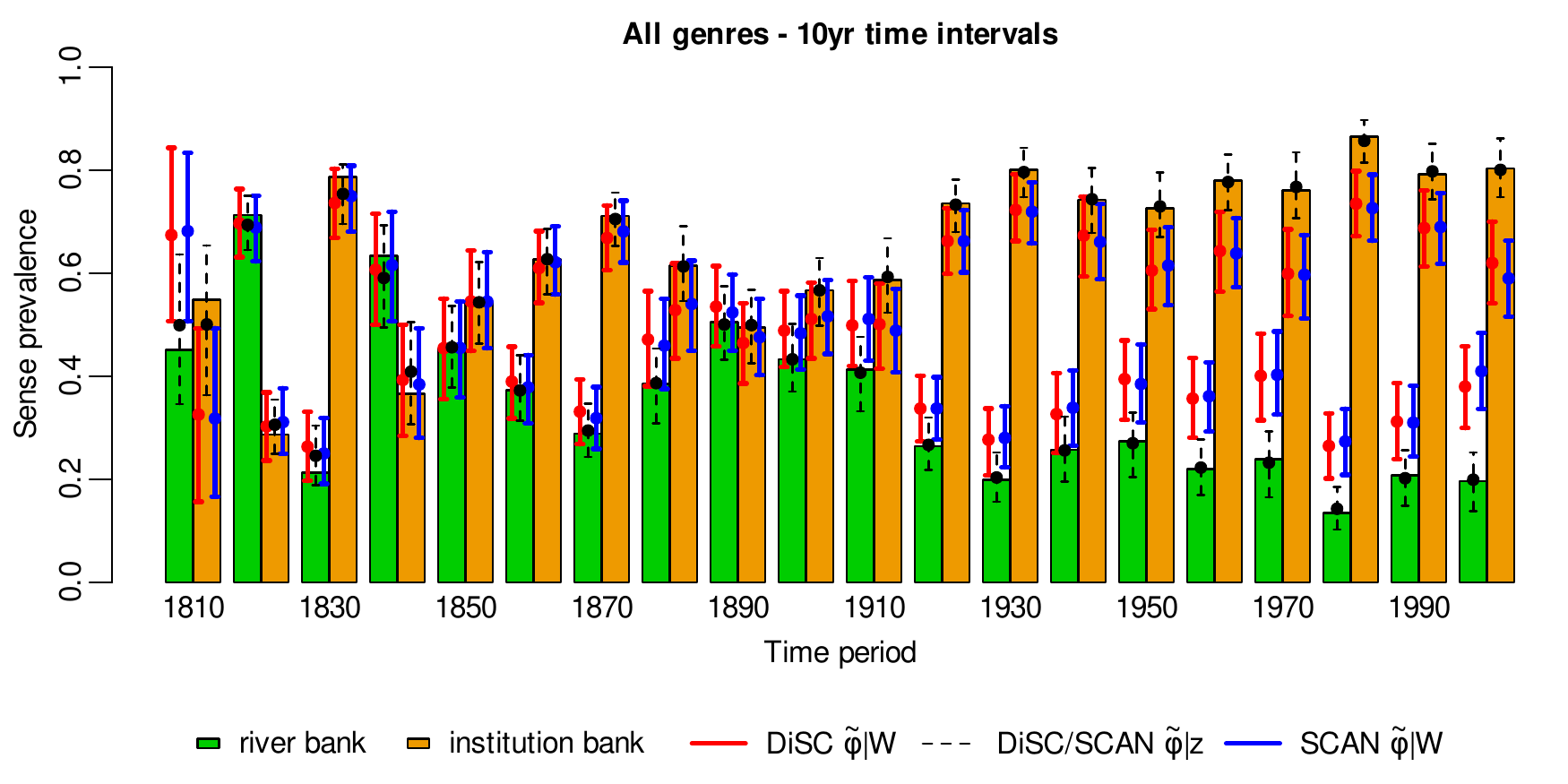}
\vspace*{-20pt}
\caption{"Bank" expert-annotated empirical sense prevalence (coloured bars with height $N_{k,g,t}^{o} / \sum_{l=1}^K N_{l,g,t}^{o}$ for each $k,g,t$) with 95\% HPD intervals (error bars) and posterior means (circles) from the model output. The first graph omits SCAN $\tilde{\phi}|W$ error bars since we could not get the MCMC to converge for SCAN.}
\label{fig:bank_thin_error_bars}
\end{figure}

We experimented with thinner time intervals on the "bank" data, as shown in Figure~\ref{fig:bank_thin_error_bars}. As expected, the uncertainty increases due to less data in each interval, but the Brier scores remain comparable to those for the 20-year intervals ($BS=0.15$ for DiSC). The 5-year graph shows an interesting and rather dramatic change in the empirical sense prevalence around 1925, which is smoothed when we take wider intervals. This sharp change could be caused by changes in the makeup of the source texts in the corpus, e.g. if a number of finance-related texts where introduced at this time. However, we did not identify a change of this sort in the corpus. We should be aware that $\tilde\phi$ (and $\tilde \psi$) measures prevalence in a language sample (i.e. the corpus) and not prevalence in historical language use itself.

In choosing an appropriate interval, two factors come into play. On the one hand, subject specialists have some understanding of the timescale of variation, and time intervals should be no greater than that scale. On the other hand, it is natural to take them as short as possible, subject to having enough data in each interval to usefully inform parameter estimates. There may be some interplay with $\kappa_\phi$, $\kappa_\psi$ and $\kappa_\theta$ as these diffusion parameters would naturally be scaled with interval length, so smaller for shorter intervals. When intervals are short, the fit becomes more sensitive to the choice of these hyperparameter priors. Using longer intervals makes the inference more robust to these choices. We encourage user exploration.

\section{Synthetic data additional results}
\label{sec:appendix:synthetic}

\begin{figure}
\includegraphics[width = 1\textwidth, keepaspectratio]{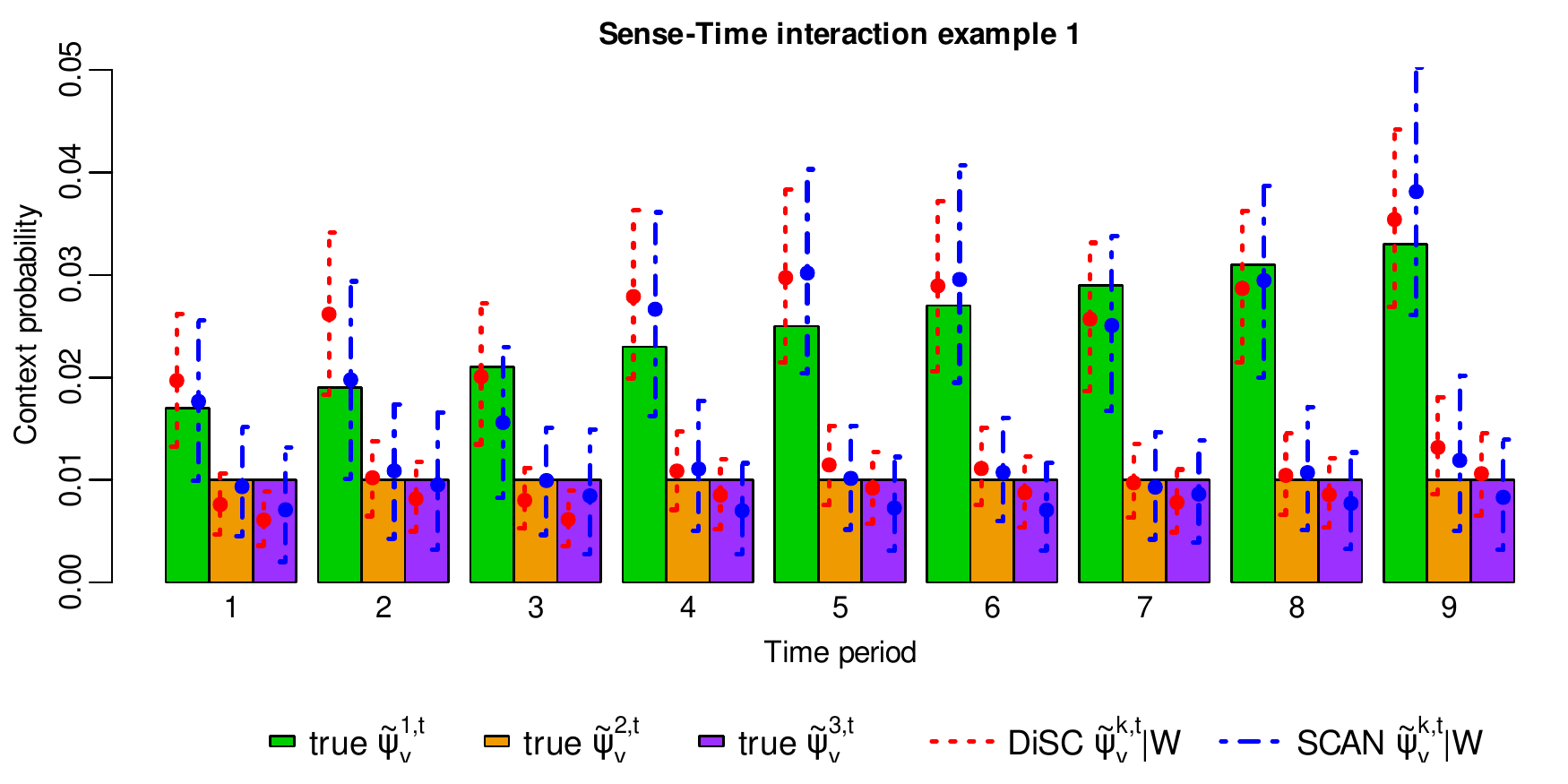}
\includegraphics[width = 1\textwidth, keepaspectratio]{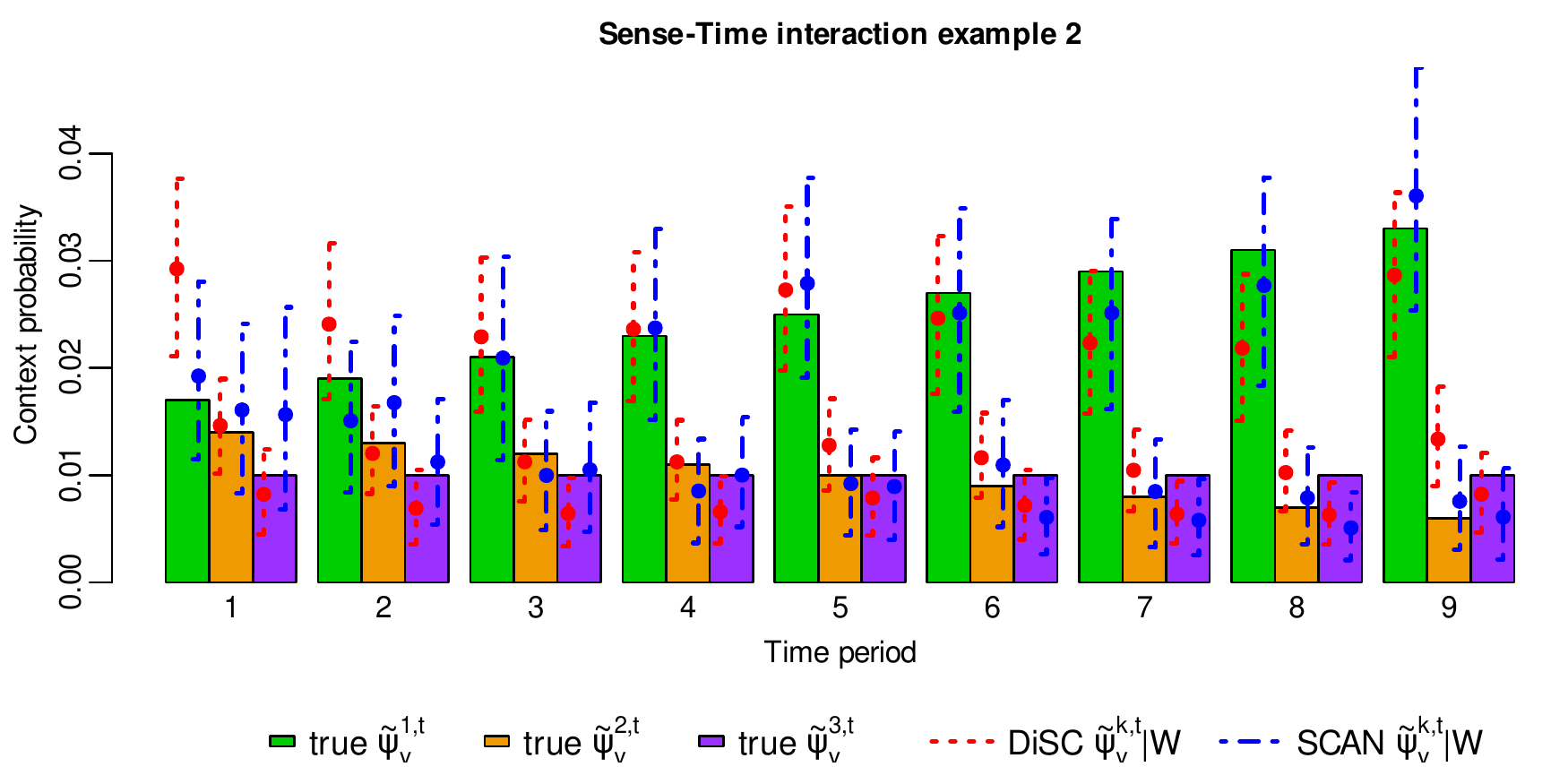}
\includegraphics[width = 1\textwidth, keepaspectratio]{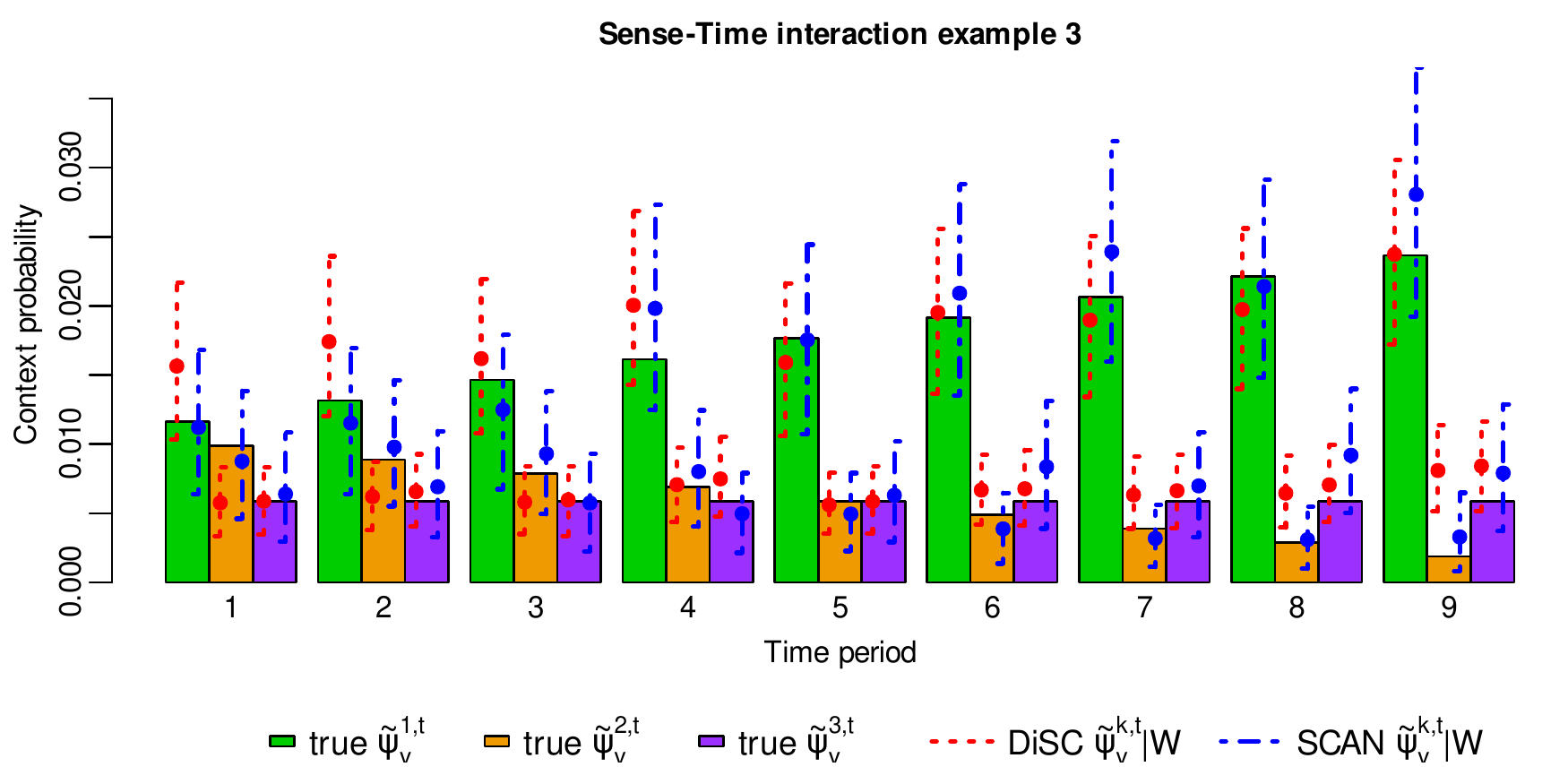}
\vspace*{-20pt}
\caption{Examples with explicit sense-time interaction, showing the context probabilities for a typical word $v$ under different senses over time. True probabilities are represented by the coloured bars, 95\% HPD intervals by the error bars, and posterior means by the circles.}
\label{fig:interaction}
\end{figure}

We constructed some examples with \textit{explicit} sense-time interaction (cf. Section~\ref{sec:experiment_synthetic}) on a small vocabulary of 100 words, so that all words have a relatively high probability of appearing in the context of any sense. 60 context words are designed to have sense-time interaction in examples 1 and 2, the other 40 being noise, whereas all 100 are designed to have sense-time interaction in example~3. For these words, the context probability $\tilde{\psi}^{k,t}_v$ either increases or decreases with time in one sense $k$, whilst doing the opposite or staying constant in the other senses. It must be emphasised that these examples are highly artificial and not reflective of real-world scenarios. The DiSC and SCAN posteriors $\tilde{\psi}_v|W$ for a single word $v$ in the three examples, together with the true $\tilde{\psi}_v$, are shown in Figure~\ref{fig:interaction}. The interaction effect is progressively stronger from example~1 to example~3. It can be seen that the DiSC posterior $\tilde{\psi}^{k,t}_v | W$ is rather flat over time for all senses in example~2, and is thus inaccurate at time 1 for the green sense and at time 9 for the orange sense. Similar inaccuracies can be seen in example~3 for the orange sense at the start and end.

The Brier scores on these examples are shown in Table~\ref{tab:interaction_brier_scores}. The performance of DiSC and SCAN on sense-labelling is comparable in all cases. Even where the interaction effect is very strong in examples~2 and~3, DiSC does not perform much worse than SCAN. Note that having more words with explicit interaction in example~3 reduces the proportion of noisy words, and hence makes it easier to identify the sense. We attempted to construct an example where DiSC fails but SCAN performs well, but were unable to do so. (In any example where DiSC failed, so did SCAN.) Moreover, it proved difficult to get SCAN to converge on these examples due to label-switching between time periods and multi-modality. It appears that not only is sense-time interaction rare in any real-world scenario, but even where it does actually exist the gains from omitting it from the model for sense-labelling purposes far outweigh the small loss in accuracy. 

\begin{table}
\caption{\label{tab:interaction_brier_scores} Brier scores on synthetic data with explicit sense-time interaction}
\centering
\begin{tabular}{l r r r}
    \toprule
         & Example 1  & Example 2  & Example 3 \\
    \midrule
    DiSC & 0.382 & 0.386 & 0.252 \\
    SCAN & 0.387 & 0.350 & 0.210 \\
    \bottomrule
\end{tabular}
\end{table}

\section{"Kosmos" additional results}
\label{sec:appendix:kosmos}

It may be argued that retaining only the snippets that an expert was able to annotate from context alone (category "collocates", cf. Section \ref{sec:data}) biases the data in the sense that the probabilities $\tilde{\phi}$ and $\tilde{\psi}$ will in general change. A related concern is that we have made the problem easier than the one we face when applying the method to data without knowing which snippets do not admit a human classification. The "kosmos" dataset contains 1,469 snippets, of which 1,144 are of the type "collocates". The remaining 325 snippets therefore represent 22\% of the data, which is a very significant level of noise given the small and sparse dataset. In our experiments, we tried to follow as closely as possible what \citet{GASC_2019arXiv190305587P} did, and therefore removed this noise in order to ensure a fair comparison with GASC. 

Recall that, in choosing the number of senses $K$ we recommend running in a semi-supervised mode: are the most frequently associated words under each sense meaningful to the user? We repeat the analysis with non-collocates included and choose $K$ on this criterion. 
We find that neither DiSC nor GASC identify recognisable meanings on $K=3$ senses, and the Brier score is higher than 0.67 (the Brier score under uniform random sense assignment).

When we run DiSC with $K=4$ senses on all the "kosmos" data, we identify three of these senses with decoration, order or world based on the most probable context words under each sense. The fourth sense has no recognisable meaning and we think of this as a sense the model is using to capture "noise". We were unable to get the GASC model to converge even with this setting --- something we found typical for GASC. 

We normalise the sense probabilities $\hat{p}(z_d=k)$ (cf. equation \eqref{eq:brier_score}) over the three recognised senses and compute the Brier score on type "collocates" data. Thus, in evaluating performance, we exclude snippets a human could not label. This gives a score of 0.38 for $K=4$ (slightly better than the 0.41 we advertised for $K=3$ senses on the data with non-collocates removed). The equivalent sense prevalence graph from the DiSC $K=4$ run is in Figure~\ref{fig:kosmos_error_bars}, and shows good agreement with the run conditioned upon the true sense labels.
For completeness, if we analyse all the data with $K=4$ senses, and compute the Brier score on all the data (i.e. not just on collocates), using the sense labels the human expert assigned to non-collocates from broader contextual considerations outside the words around the target word, we get a score higher than 0.67. Thus, in situations where a human cannot annotate the sense based on context, our model cannot either.

\begin{figure}[!t]
\centering
\includegraphics[width = 1\textwidth, keepaspectratio]{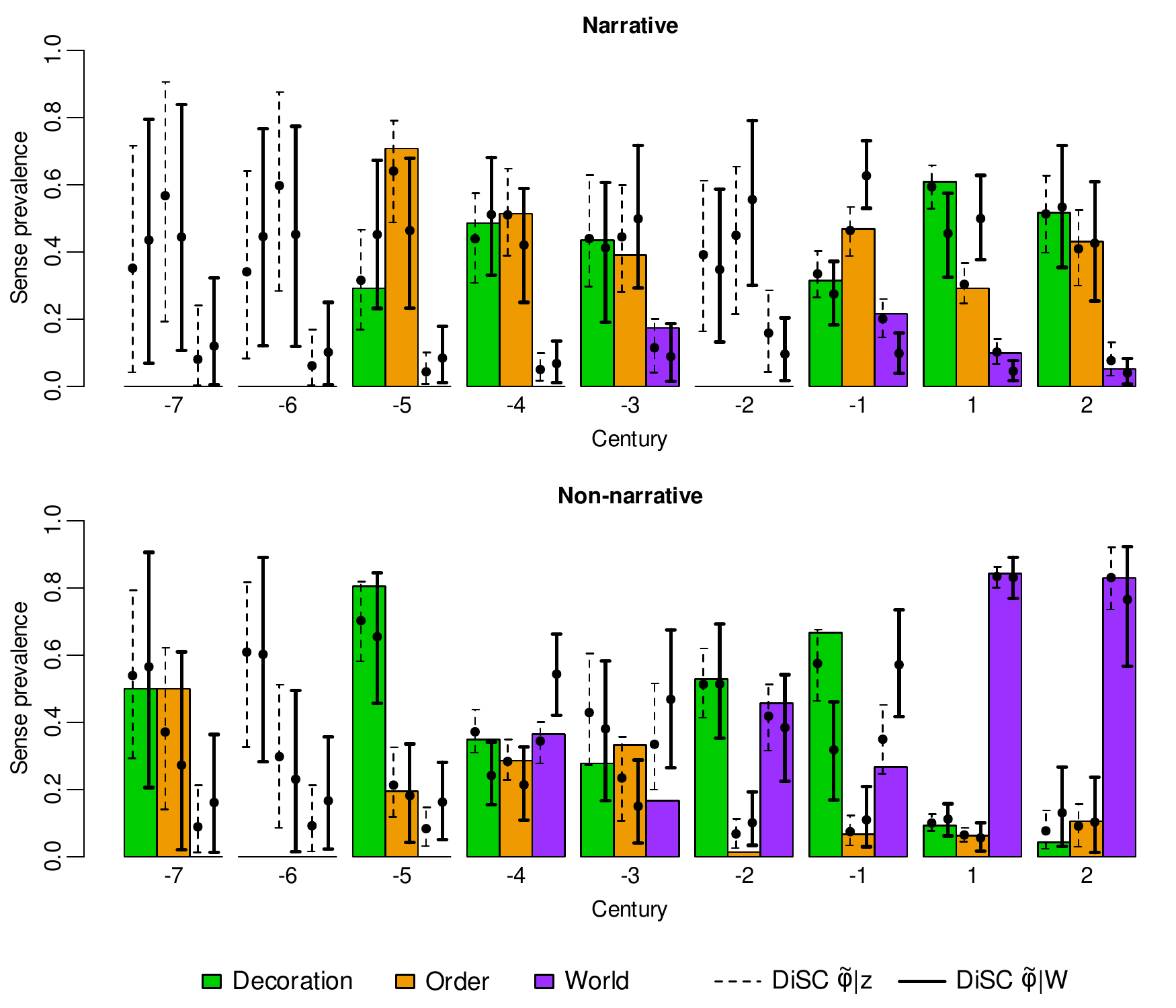}
\vspace*{-20pt}
\caption{"Kosmos" expert-annotated empirical sense prevalence (coloured bars with height $N_{k,g,t}^{o} / \sum_{l=1}^{3} N_{l,g,t}^{o}$ for each $k,g,t$) with 95\% HPD intervals (error bars) and posterior means (circles) from the $K=4$ DiSC run}
\label{fig:kosmos_error_bars}
\end{figure}

\end{appendices}

\bibliographystyle{rss}
\bibliography{bibliography.bib}

\end{document}